\pdfoutput=1

\documentclass[11pt]{article}

\usepackage[]{acl}

\usepackage{times}
\usepackage{latexsym}

\usepackage[T1]{fontenc}

\usepackage[utf8]{inputenc}

\usepackage{microtype}

\title{Accurate Online Posterior Alignments for Principled Lexically-Constrained Decoding}
\author{Soumya Chatterjee \\ IIT Bombay \\ \texttt{soumya@cse.iitb.ac.in}
        \And
        Sunita Sarawagi \\ IIT Bombay \\ \texttt{sunita@iitb.ac.in}
        \And
        Preethi Jyothi \\ IIT Bombay \\ \texttt{pjyothi@cse.iitb.ac.in}
}

\usepackage[textsize=small]{todonotes}

\usepackage{amsmath}
\usepackage{amssymb}
\usepackage[mode=buildnew]{standalone}
\usepackage{multirow}
\usepackage{color}
\usepackage[normalem]{ulem}
\usepackage{adjustbox}

\DeclareMathOperator{\argmax}{argmax}
\newcommand{\mb}{\mathbf}
\newcommand{\mcal}[1]{\mathcal{#1}}
\newcommand{\calA}{\mcal{A}}
\newcommand{\R}{\mathbb{R}}

\newcommand{\encOut}{\mb H}
\newcommand{\keys}{\mb K}
\newcommand{\values}{\mb V}

\newcommand{\qvector}{\mb q}
\newcommand{\decSAOut}{\mb g}

\newcommand{\wordemb}{\mb e}
\newcommand{\postattsub}{\text{post}}
\newcommand{\nheads}{\eta}
\newcommand{\apost}[1]{a^{\postattsub}_{#1}}
\newcommand{\apostvec}[1]{\mb a^{\postattsub}_{#1}}

\newcommand{\shiftatt}{\textsc{ShiftATT}}
\newcommand{\shiftaet}{\textsc{ShiftAET}}
\newcommand{\prioratt}{\textsc{PriorATT}}

\newcommand{\naiveatt}{\textsc{NaiveATT}}

\newcommand{\alignvdba}{Align-VDBA}

\newcommand{\postatt}{\textsc{PostAln}}

\usepackage{algorithm,algorithmicx}
\usepackage[noend]{algpseudocode}

\usepackage{enumitem}

\newcommand{\h}[1]{\textbf{#1}}

\begin{document}
\maketitle
\begin{abstract}
Online alignment in machine translation refers to the task of aligning a target word to a source word when the target sequence has only been partially decoded. 
Good online alignments facilitate important applications such as lexically constrained translation where user-defined dictionaries are used to inject lexical constraints into the translation model. We propose a novel posterior alignment technique that is truly online in its execution and superior in terms of alignment error rates compared to existing
methods. Our proposed inference technique jointly considers alignment and token probabilities in a principled manner and can be seamlessly integrated within existing constrained beam-search decoding algorithms. On five language pairs, including two distant language pairs, we achieve consistent drop in alignment error rates. When deployed on seven lexically constrained translation tasks, we achieve significant improvements in BLEU specifically around the constrained positions.
\end{abstract}

\section{Introduction}
Online alignment seeks to align a target word to a source word  at the decoding step when the word is output in an auto-regressive neural translation model~\citep{kalchbrenner-blunsom-2013-recurrent, cho-etal-2014-properties, sutskever-etal-2014-sequence}. This is  unlike the more popular offline alignment task that uses the entire target sentence~\citep{och-ney-2003-systematic}. State of the art methods of offline alignment based on matching of whole source and target sentences \cite{jalili-sabet-etal-2020-simalign,dou-neubig-2021-word} are not applicable for online alignment where we need to commit on the alignment of a target word based on only the generated prefix thus far. 

An important application of online alignment is lexically constrained translation which allows injection of domain-specific terminology and other phrasal constraints during decoding~\cite{hasler-etal-2018-neural, hokamp-liu-2017-lexically, alkhouli-etal-2018-alignment, crego-etal-2016-systrans}. 
Other applications include preservation of markups between the source and target~\cite{muller-2017-treatment}, and supporting source word edits in summarization~\cite{shen-etal-2019-improving}. These applications need to infer the specific source token which aligns with output token. Thus, alignment and translation is to be done simultaneously.

Existing online alignment methods can be categorized into Prior and Posterior alignment methods.   Prior alignment methods~\cite{garg-etal-2019-jointly,song-etal-2020-alignment} extract alignment based on the attention at time step $t$ when outputting token $y_t$.  The attention probabilities at time-step $t$ are conditioned on tokens output before time $t$. Thus, the alignment is estimated \emph{prior} to observing $y_t$.  Naturally, the quality of alignment can be improved if we condition on the target token $y_t$~\cite{shankar2018posterior}. This motivated \citet{chen-etal-2020-accurate} to propose a posterior alignment method where alignment is calculated from the attention probabilities at the next decoder step $t+1$.  While alignment quality improved as a result, their method is not truly online since it does not generate alignment  \emph{synchronously} with the token. The delay of one step makes it difficult and cumbersome to incorporate terminology constraints during beam decoding.

We propose a truly online posterior alignment method that provides higher alignment accuracy than existing online methods, while also being synchronous. Because of that we can easily integrate posterior alignment to improve lexicon-constrained translation in state of the art constrained beam-search algorithms such as VDBA~\cite{hu-etal-2019-improved}.
Our method (\alignvdba) presents a significant departure from existing papers on alignment-guided constrained translation~\cite{chen-etal-2020-accurate,song-etal-2020-alignment} that employ a greedy algorithm with poor constraint satisfaction rate (CSR). For example, on a ja$\rightarrow$en their CSR is 20 points lower than ours.  
Moreover, the latter does not benefit from larger beam sizes unlike VDBA-based methods that significantly improve with larger beam widths. Compared to~\citet{chen-etal-2020-accurate}, our method improves average overall BLEU scores by 1.2 points and average BLEU scores around the constrained span by up to 9 points.  In the evaluations performed in these earlier work, VDBA was not allocated the slightly higher beam size needed to pro-actively enforce constraints without compromising BLEU. 
Compared to \citet{hu-etal-2019-improved} (VDBA), this paper's contributions include online alignments and their use in more fluent constraint placement and efficient allocation of beams.

\paragraph{Contributions}
\begin{itemize}[leftmargin=1em, itemsep=-0.4em, topsep=0em]
    \item A truly online posterior alignment method that integrates into existing NMT sytems via a trainable light-weight module.
    \item Higher online alignment accuracy on five language pairs including two distant language pairs where we improve over the best existing method in seven out of ten translation tasks.
    \item Principled method of modifying VDBA to incorporate posterior alignment probabilities in lexically-constrained decoding. VDBA enforces constraints ignoring source alignments; our change (\alignvdba) leads to more fluent constraint placement and significant BLEU increase particularly for smaller beams.
    \item Establishing that VDBA-based pro-active constrained inference should be preferred over prevailing greedy alignment-guided inference~\cite{chen-etal-2021-lexically,song-etal-2020-alignment}. 
    Further, VDBA and our \alignvdba\ inference with beam size 10 provide 1.2 BLEU increase over  these methods with the same beam size.
\end{itemize}

\section{Posterior Online Alignment}
Given a sentence $\mb x = x_1,\ldots, x_S$ in the source language and a sentence $\mb y = y_1, \ldots, y_T$ in the target language, an alignment $\calA$ between the word strings is a subset of the Cartesian product of the word positions \citep{brown-etal-1993-mathematics, och-ney-2003-systematic}: $\calA \subseteq \{(s,t): s = 1,\ldots,S ;\, t=1,\ldots,T\}$ such that the aligned words can be considered translations of each other. 
An online alignment at time-step $t$ commits on alignment of the $t$\textsuperscript{th} output token  conditioned only on $\mb x$ and $\mb y_{< t} = y_1, y_2,\ldots y_{t-1}$. Additionally, if token $y_t$ is also available we call it a posterior online alignment. 
We seek to embed online alignment with existing NMT systems.  We will first briefly describe the architecture of state of the art NMT systems. We will then elaborate on how alignments are computed from attention distributions in prior work and highlight some limitations, before describing our proposed approach.

\subsection{Background}
Transformers~\cite{vaswani-etal-2017-attention} adopt the popular encoder-decoder paradigm used for sequence-to-sequence modeling~\cite{cho-etal-2014-properties, sutskever-etal-2014-sequence, bahdanau-etal-2015-neural}. The encoder and decoder are both multi-layered networks with each layer consisting of a multi-headed self-attention and a feedforward module. The decoder layers additionally use multi-headed attention to encoder states.  We elaborate on this mechanism next since it plays an important role in alignments.

\subsubsection{Decoder-Encoder Attention in NMTs}
The encoder transforms the $S$ input tokens into a sequence of token representations $\encOut \in \R^{S\times d}$. Each decoder layer (indexed by $\ell \in \{1,\ldots,L\}$) computes multi-head attention over $\encOut$ by aggregating outputs from a set of $\nheads$ independent attention heads. The attention output from a single head $n \in \{1,\ldots,\nheads\}$ in decoder layer $\ell$ is computed as follows. Let the output of the self-attention sub-layer in decoder layer $\ell$ at the $t^\text{th}$ target token be denoted as $\decSAOut^{\ell}_t$. Using three projection matrices $\mb W^{\ell, n}_Q$, $\mb W^{\ell, n}_V$, $\mb W^{\ell, n}_K \in  \R^{d\times d_n}$, the query vector $\qvector_t^{\ell, n} \in \R^{1\times d_n}$ and key and value matrices, $\keys^{\ell, n} \in \R^{S\times d_n}$ and $\values^{\ell, n} \in \R^{S\times d_n}$, are computed using the following projections: $\qvector^{\ell, n}_t = \decSAOut^{\ell}_t \mb W^{\ell, n}_Q$, $\keys^{\ell, n} = \encOut \mb W^{\ell, n}_K$, and $\values^{\ell, n} = \encOut \mb W^{\ell, n}_V$.%
\footnote{$d_n$ is typically set to $\frac{d}{\eta}$ so that a multi-head attention layer does not introduce more parameters compared to a single head attention layer.}
These are used to calculate the attention output from head $n$, $\mb Z_t^{\ell, n} = P(\mb a_t^{\ell,n}| \mb x, \mb y_{< t}) \values^{\ell, n}$, where:
\begin{equation}
    P(\mb a_t^{\ell,n}| \mb x, \mb y_{< t}) = \mathrm{softmax}\left(\frac{\qvector_t^{\ell, n} (\keys^{\ell, n})^\intercal}{\sqrt{d}}\right)
    \label{eqn:attention}
\end{equation}
For brevity, the conditioning on $\mb x, \mb y_{<t}$ is dropped and $P(\mb a_t^{\ell,n})$ is used to refer to $P(\mb a_t^{\ell,n}| \mb x, \mb y_{< t})$ in the following sections.

Finally, the multi-head attention output is given by $[\mb Z_t^{\ell, 1}, \ldots, \mb Z_t^{\ell, \eta}]\mb W^O$ where $[\ ]$ denotes the column-wise concatenation of matrices and $\mb W^O \in \R^{d\times d}$ is an output projection matrix. 

\subsubsection{Alignments from Attention}
Several prior work have proposed to extract word alignments from the above attention probabilities. For example \citet{garg-etal-2019-jointly} propose a simple method called \naiveatt\ that
aligns a source word to the  $t^{\text{th}}$ target token using $ \displaystyle \argmax_{j} \frac{1}{\eta} \sum_{n=1}^{\eta} P(a^{\ell, n}_{t,j} | \mb x, \mb y_{< t})$ where $j$ indexes the source tokens. 
In \naiveatt, we note that the attention probabilities $P(a^{\ell, n}_{t,j}|\mb x,\mb y_{< t})$ at decoding step $t$ are not conditioned on the current output token $y_t$. Alignment quality would benefit from conditioning on $y_t$ as well. This observation prompted \citet{chen-etal-2020-accurate} to extract alignment of token $y_t$ using attention $P(a^{\ell, n}_{t,j} | \mb x, \mb y_{\le t})$ computed at time step $t+1$. The asynchronicity inherent to this shift-by-one approach (\shiftatt) makes it difficult and more computationally expensive to incorporate lexical constraints during beam decoding. 

\subsection{Our Proposed Method: \postatt}
\label{sec:postatt}

\begin{figure}[t]
    \centering
    \includegraphics[width=0.9\linewidth]{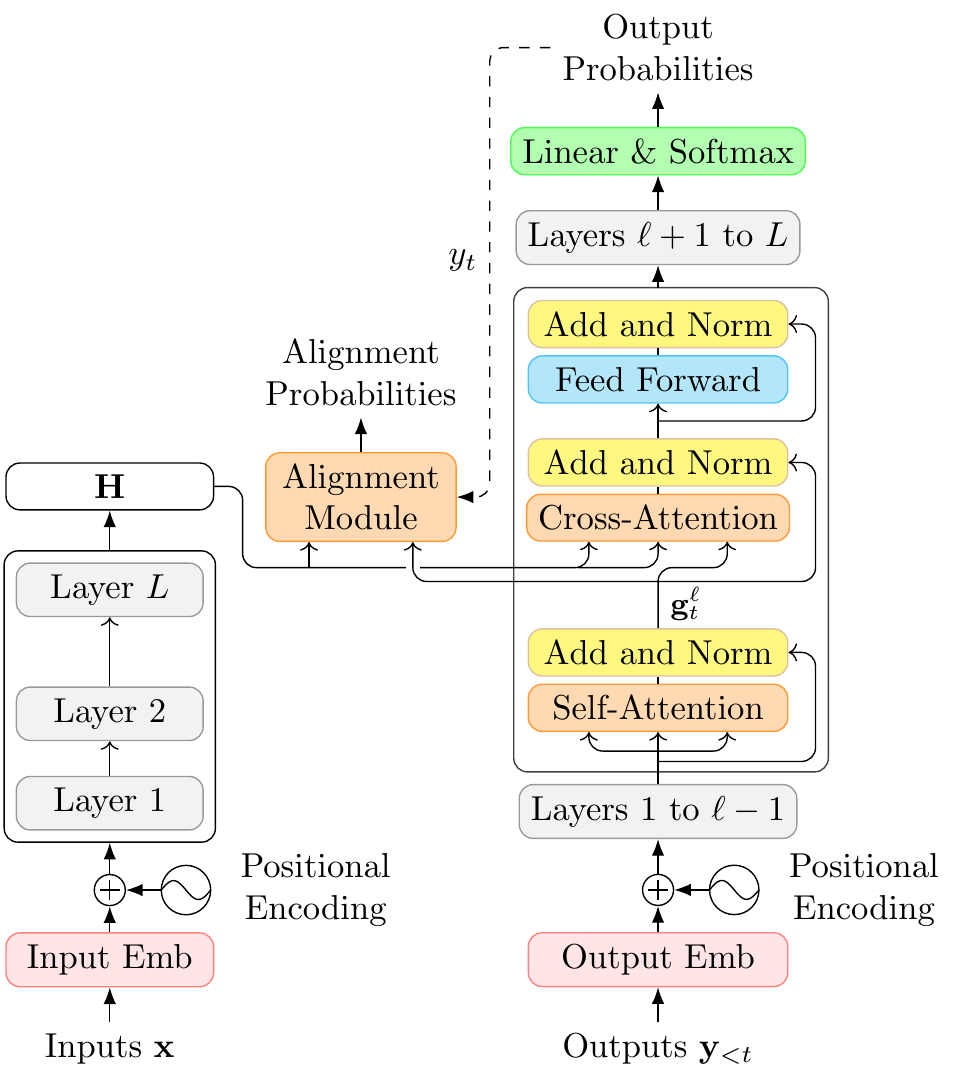}
    \caption{Our alignment module is an encoder-decoder attention sub-layer, similar to the existing cross-attention sub-layer. It takes as inputs the encoder output $\encOut$ as the key, and the concatenation of the output of the previous self-attention layer $\decSAOut^\ell_t$ and the currently decoded token $y_t$ as the query, and outputs posterior alignment probabilities $\apostvec{t}$.}
    \label{fig:alignment-module}
\end{figure}

We propose \postatt\ that produces posterior alignments synchronously with the output tokens, while being more computationally efficient compared to previous approaches like \shiftatt. We incorporate a lightweight alignment module to convert prior attention to posterior alignments in the same decoding step as the output. Figure~\ref{fig:alignment-module} illustrates how this alignment module fits within the standard Transformer architecture.

The alignment module is placed at the penultimate decoder layer $\ell=L-1$ and takes as input (1)~the encoder output $\encOut$, (2)~the output of the self-attention sub-layer of decoder layer $\ell$, $\decSAOut^\ell_t$ and, (3)~the embedding of the decoded token $\wordemb(y_t)$.   Like in standard attention it projects  $\encOut$ to obtain a key matrix, but to obtain the query matrix it uses both decoder state $\decSAOut^\ell_t$ (that summarizes $\mb y_{<t}$) and $\wordemb(y_t)$ to compute the posterior alignment  $P(\apostvec{t})$ as: 
\begin{align*}
    &P(\apostvec{t}) = \frac{1}{\eta}\sum_{n=1}^\eta\mathrm{softmax}\left(\frac{\qvector_{t, \postattsub}^{n} (\keys_{\postattsub}^{n})^\intercal}{\sqrt{d}}\right),\  \\
    &\qvector^{n}_{t, \postattsub} = [\decSAOut^{\ell}_t, \wordemb(y_t)] \mb W^{n}_{Q, \postattsub},\ \keys^{n}_{\postattsub} = \encOut \mb W^{n}_{K, \postattsub}
\end{align*}
Here $\mb W^{n}_{Q, \postattsub} \in \R^{2d\times d_n}$ and $\mb W^{n}_{K, \postattsub} \in \R^{d\times d_n}$.

This computation is synchronous with producing the target token $y_t$, thus making it compatible with beam search decoding (as elaborated further in Section~\ref{sec:translation}). It also accrues minimal computational overhead since $P(\apostvec{t})$ is defined using $\encOut$ and $\decSAOut^{L-1}_t$, that are both already cached during a standard decoding pass.
Note that if the query vector $\qvector^{n}_{t,\postattsub}$ is computed using only $\decSAOut^{L-1}_t$, without concatenating $\wordemb(y_t)$, then we get prior alignments that we refer to as \prioratt. In our experiments, we explicitly compare \prioratt\ with \postatt\ to show the benefits of using  $y_t$ in deriving alignments while keeping the rest of the architecture intact. 
 
\paragraph*{Training}
Our posterior alignment sub-layer is trained using alignment supervision, while freezing the rest of the translation model parameters. Specifically, we train a total of $3d^2$ additional parameters across the matrices $\mb W^{n}_{K, \postattsub}$ and $\mb W^{n}_{Q, \postattsub}$.
Since gold alignments are very tedious and expensive to create for large training datasets, alignment labels are typically obtained using existing techniques. We use bidirectional symmetrized \shiftatt\ alignments, denoted by $S_{i,j}$ that refers to an alignment between the $i^{\text{th}}$ target word and the $j^{\text{th}}$ source word, as reference labels to train our alignment sub-layer.
Then the objective (following~\citet{garg-etal-2019-jointly}) can be defined as:
\begin{equation*}
    \max_{\mb W^{n}_{Q, \postattsub},\mb W^{n}_{K, \postattsub}} \frac{1}{T} \sum_{i=1}^{T} \sum_{j=1}^{S} S_{i,j} \log(P(\apost{{i, j}}|\mb x,\mb y_{\le i}))
\end{equation*}

Next, we demonstrate the role of posterior online alignments on an important downstream task.

\section{Lexicon Constrained Translation}
\label{sec:translation}

In the lexicon constrained translation task, for each to-be-translated sentence $\mb x$, we are given a set of source text spans and the corresponding target tokens in the translation. A constraint $\mcal{C}_j$ comprises a pair $(\mcal{C}^x_j, \mcal{C}^y_j)$ where $\mcal{C}^x_j = (p_j, p_j+1 \ldots, p_j+\ell_j)$ indicates input token positions, and $\mcal{C}^y_j = ( y^j_{1}, y^j_{2} \ldots,  y^j_{m_j})$ denote target tokens that are translations of the input tokens $x_{p_j} \ldots x_{p_j+\ell_j}$. For the output tokens we do not know their positions in the target sentence. The different constraints are non-overlapping and each is expected to be used exactly once.
The goal is to translate the given sentence $\mb x$ and satisfy as many constraints in $\mcal{C} = \bigcup_j \mcal{C}_j$ as possible while ensuring fluent and correct translations.
Since the constraints do not specify target token position, it is natural to use online alignments to guide when a particular constraint is to be enforced.

\subsection{Background: Constrained Decoding}
\label{sec:translation-background}
Existing inference algorithms for incorporating lexicon constraints differ in how pro-actively they enforce the constraints. A passive method is used in \citet{song-etal-2020-alignment} where constraints are enforced only when the prior alignment is at a constrained source span.
Specifically, if at decoding step $t$, $i = \argmax_{i'} P(a_{t,i'})$ is present in some constraint $\mcal{C}^x_j$, the output token is fixed to the first token $y^j_{1}$ from $\mcal{C}^y_j$. Otherwise, the decoding proceeds as usual. Also, if the translation of a constraint $\mcal{C}_j$ has started, the same is completed ($y^j_{2}$ through $ y^j_{m_j}$) for the next $m_j-1$ decoding steps before resuming unconstrained beam search. The pseudocode for this method is provided in Appendix~\ref{appendix:algo2}.

For the posterior alignment methods of \citet{chen-etal-2020-accurate} this leads to a rather cumbersome inference \citep{chen-etal-2021-lexically}.  First, at step $t$ they predict a token $\hat{y_t}$, then start decoding step $t+1$ with $\hat{y_t}$ as input to compute the posterior alignment from attention at step $t+1$. If the maximum alignment is to the constrained source span $\mcal{C}^x_j$ they {\em revise} the output token to be ${y^j_1}$ from $\mcal{C}^y_j$, but the output score for further beam-search continues to be of $\hat{y_t}$.
In this process both the posterior alignment and token probabilities are misrepresented since they are both based on $\hat{y_t}$ instead of the finally output token ${y^j_1}$.  The decoding step at $t+1$ needs to be restarted after the revision. The overall algorithm continues to be normal beam-search, which implies that the constraints are not enforced pro-actively.  
\begin{algorithm*}[t]
\begin{small}
  \caption{\alignvdba: Modifications to DBA shown in blue. (Adapted from \citet{post-vilar-2018-fast})}
  \begin{algorithmic}[1]
  \State {\bf Inputs} beam: $K$ hypothesis in beam, scores: $K \times |V_T|$ matrix of scores where scores$[k,y]$ denotes the score of $k$\textsuperscript{th} hypothesis extended with token $y$ at this step, constraints: $\{(\mcal{C}^x_j, \mcal{C}^y_j)\}$, threshold
  \State candidates $\gets$ [$(k,y,$ scores[$k,y$], beam[$k$].constraints.add($y$)] for $k,y$ in \Call{argmax\_K}{scores}
  \For{$1 \leq k \leq K$} \Comment{Go over current beam}
    \ForAll{$y \in V_T$ that are unmet constraints for beam[$k$]} \Comment{Expand new constraints}
      \State {\color{darkblue} alignProb $\gets$ $\Sigma_{\text{constraint\_xs}(y)}$ \Call{\postatt}{$k, y$}} \label{alg-step:mod1} \Comment{Modification in blue (Eqn~\eqref{eqn:joint-prob})} 
      \If{{\color{darkblue} alignProb > threshold}} \label{alg-step:mod3}
          \State {\color{darkblue} candidates.append( $(k,y,$ scores[$k,y$] $\times$ alignProb), beam[$k$].constraints.add($y$) ) )} \label{alg-step:mod2}
      \EndIf
      \State {\color{black!20}candidates.append( $(k,y,$ scores[$k,y$], beam[$k$].constraints.add($y$) ) )} \Comment{Original DBA Alg.}
    \EndFor
    \State $w$ = \Call{argmax}{scores[$k, :$]}
    \State candidates.append( ($k, w$, scores[$k, w$], beam[$k$].constraints.add($w$) ) ) \Comment{Best single word}
  \EndFor
  \State newBeam $\gets$ \Call{allocate}{candidates, $K$} \label{alg-step:allocate}
  \end{algorithmic}
  \label{alg:vdba}
\end{small}
\end{algorithm*}

Many prior methods have proposed more pro-active methods of enforcing constraints, including
the Grid Beam Search (GBA,~\citet{hokamp-liu-2017-lexically}), Dynamic Beam Allocation (DBA,~\citet{post-vilar-2018-fast}) and Vectorized Dynamic Beam Allocation (VDBA,~\citet{hu-etal-2019-improved}). The latest of these, VDBA, is efficient and available in public NMT systems~\cite{ott-etal-2019-fairseq, hieber-etal-2020-sockeye}.   Here multiple \textit{banks},  each corresponding to a particular number of completed constraints, are maintained. At each decoding step, a hypothesis can either start a new constraint and move to a new bank or continue in the same bank (either by not starting a constraint or progressing on a constraint mid-completion). This allows them to achieve near 100\% enforcement.
However, VDBA enforces the constraints by considering only the target tokens of the lexicon and totally ignores the alignment of these tokens to the source span.  This could lead to constraints being placed at unnatural locations leading to loss of fluency.  Examples appear in Table~\ref{tab:anec} where we find that VDBA just attaches the constrained tokens at the end of the sentence. 

\subsection{Our Proposal: \alignvdba}
We modify VDBA with alignment probabilities to better guide constraint placement.  The score of a constrained token is now the joint probability of the token, and the probability of the token being aligned with the corresponding constrained source span. Formally, if the current token $y_t$ is a part of the $j$\textsuperscript{th} constraint \textit{i.e.} $y_t \in \mcal{C}^y_j$, the generation probability of $y_t$, $P(y_t | \mb x, \mb y_{<t})$ is scaled by multiplying with the alignment probabilities of $ y_t$ with $\mcal{C}^x_j$, the source span for constraint $i$. Thus, the updated probability is given by:
\begin{equation}
\small
\underbrace{P(y_t, \mcal{C}^x_j | \mb x, \mb y_{<t})}_\text{Joint Prob} = \underbrace{P(y_t | \mb x, \mb y_{<t})}_\text{Token Prob} \underbrace{\sum_{r \in \mcal{C}^x_j} P(\apost{{t, r}} | \mb x, \mb y_{\le t} )}_\text{Src Align. Prob.}
\label{eqn:joint-prob}
\end{equation}
$P(y_t, \mcal{C}^x_j | \mb x, \mb y_{<t})$ denotes the joint probability of outputting the constrained token and the alignment being on the corresponding source span. 
Since the supervision for the alignment probabilities was noisy, we found it useful to recalibrate the alignment distribution using  a temperature scale $T$, so that the recalibrated probability is $\propto \Pr(\apost{{t, r}} | \mb x, \mb y_{\le t} )^{\frac{1}{T}}$. We used $T=2$ i.e., square-root of the alignment probability.

\alignvdba\ also uses posterior alignment probabilities to also improve the efficiency of VDBA. Currently, VDBA attempts beam allocation for each unmet constraint since it has no way to discriminate.  In \alignvdba\ we allocate only when the alignment probability is greater than a threshold. When the beam size is small (say 5) this yields higher accuracy due to more efficient beam utilization. We used a threshold of 0.1 for all language pairs other than ro$\rightarrow$en for which a threshold of 0.3 was used. Further, the thresholds were used for the smaller beam size of 5 and not for larger beam sizes of 10 and 20.

We present the pseudocode of our modification (steps~\ref{alg-step:mod1}, \ref{alg-step:mod3} and \ref{alg-step:mod2}, in blue) to DBA in Algorithm~\ref{alg:vdba}. Other details of the algorithm including the handling of constraints and the allocation steps (step \ref{alg-step:allocate}) are involved and we refer the reader to   \citet{post-vilar-2018-fast} and \citet{hu-etal-2019-improved} to understand these details.  The point of this code is to show that our proposed posterior alignment method can be easily incorporated into these algorithms so as to provide a more principled scoring of constrained hypothesis in a beam than the ad hoc revision-based method of \citet{chen-etal-2021-lexically}. Additionally, posterior alignments lead to better placement of constraints than in the original VDBA algorithm.

\section{Experiments}
\label{sec:expt}
We first compare our proposed posterior online alignment method on quality of alignment against existing methods in Section~\ref{expt:align}, and in Section~\ref{expt:translation}, we demonstrate the impact of the improved alignment on the lexicon-constrained translation task. 

\subsection{Setup}
\label{sec:setup}
We deploy the \texttt{fairseq} toolkit \citep{ott-etal-2019-fairseq} and use \texttt{transformer\_iwslt\_de\_en} pre-configured model for all our experiments. Other configuration parameters include: Adam optimizer with $\beta_1 = 0.9,\ \beta_2 = 0.98$, a learning rate of $5\mathrm{e}{-4}$ with 4000 warm-up steps, an inverse square root schedule, weight decay of $1\mathrm{e}{-4}$, label smoothing of $0.1$, $0.3$ probability dropout and a batch size of 4500 tokens. The transformer models are trained for 50,000 iterations. Then, the alignment module is trained for 10,000 iterations, keeping the other model parameters fixed. A joint byte pair encoding (BPE) is learned for the source and the target languages with 10k merge operation \citep{sennrich-etal-2016-neural} using \texttt{subword-nmt}. %

All experiments were done on a single 11GB Nvidia GeForce RTX 2080 Ti GPU on a machine with 64 core Intel Xeon CPU and 755 GB memory. The vanilla Transformer models take between 15 to 20 hours to train for different datasets. Starting from the alignments extracted from these models, the \postatt\ alignment module trains in about 3 to 6 hours depending on the dataset.
\begin{table}[t!]
    \centering
    \resizebox{\linewidth}{!}{
    \begin{tabular}{|l|rrr|rr|}
        \hline
         & de-en & en-fr & ro-en & en-hi & ja-en \\
        \hline
        Training & 1.9M & 1.1M & 0.5M & 1.6M & 0.3M\\
        Validation & 994 & 1000 & 999 & 25 & 1166 \\
        Test & 508 & 447 & 248 & 140 & 1235 \\
        \hline
    \end{tabular}
    }
    \caption{Number of sentence pairs for the five datasets used. Note that gold alignments are available only for the handful of sentence pairs in the test set.}
    \label{tab:data}
\end{table}
\begin{table*}[ht!]
\centering
\resizebox{\textwidth}{!}{
    \begin{tabular}{l|c|rr|rr|rr|rr|rr}
          & \multirow{2}{*}{\rotatebox{45}{Delay}} & \multicolumn{2}{c|}{de-en} & \multicolumn{2}{c|}{en-fr} & \multicolumn{2}{c|}{ro-en} & \multicolumn{2}{c|}{en-hi} & \multicolumn{2}{c}{ja-en} \\
    \cline{3-12} 
    Method &  & de$\rightarrow$en & en$\rightarrow$de & en$\rightarrow$fr & fr$\rightarrow$en & ro$\rightarrow$en & en$\rightarrow$ro & en$\rightarrow$hi & hi$\rightarrow$en & ja$\rightarrow$en & en$\rightarrow$ja \\ 
    \hline \hline
    \multicolumn{12}{c}{Statistical Methods (Not Online)} \\
    \hline
    GIZA++~\citep{och-ney-2003-systematic}     & End & 18.9 & 19.7 &  7.3 &  7.0 & 27.6 & 28.3 & 35.9 & 36.4 & 41.8 & 39.0\\
    FastAlign~\citep{dyer-etal-2013-simple}    & End & 28.4 & 32.0 & 16.4 & 15.9 & 33.8 & 35.5 &   -  &   -  &   -  &   - \\
    \hline
    \multicolumn{12}{c}{No Alignment Training} \\
    \hline
    \naiveatt~\citep{garg-etal-2019-jointly}    &  0 & 32.4 & 40.0 & 24.0 & 31.2 & 37.3 & 33.2 & 49.1 & 53.8 & 62.2 & 63.5\\
    \shiftatt~\cite{chen-etal-2020-accurate}    & +1 & 20.0 & 22.9 & 14.7 & 20.4 & 26.9 & 27.4 & 35.3 & 38.6 & 53.6 & 48.6\\
    \hline
    \multicolumn{12}{c}{With Alignment Training} \\
    \hline
    \prioratt                                   &  0 & 23.4 & 25.8 & 14.0 & 16.6 & 29.3 & 27.2 & 36.4 & 35.1 & 52.7 & 50.9\\
    \shiftaet~\cite{chen-etal-2020-accurate}    & +1 & 15.8 & \textbf{19.5} & 10.3 & \textbf{10.4} & 22.4 & 23.7 & 29.3 & 29.3 & 42.5 & \textbf{41.9}\\
    \postatt~[Ours]                             &  0 & \textbf{15.5} & \textbf{19.5} &  \textbf{9.9} & \textbf{10.4} & \textbf{21.8} & \textbf{23.2} & \textbf{28.7} & \textbf{28.9} & \textbf{41.2} & 42.2    \end{tabular}
}
\caption{AER for de-en, en-fr, ro-en, en-hi, ja-en language pairs. ``Delay" indicates the decoding step at which the alignment of the target token is available.  \naiveatt, \prioratt\ and \postatt\ are truly online and output alignment at the same time step (delay=0), while \shiftatt\ and \shiftaet\ output one decoding step later.}
\label{tab:aer-main}
\end{table*}

\subsection{Alignment Task}
\label{expt:align}
We evaluate online alignments on ten translation tasks spanning five language pairs. Three of these are popular in alignment papers~\cite{zenkel-etal-2019-adding}:
German-English (de-en), English-French (en-fr), Romanian-English (ro-en).  These are all European languages that follow the same 
subject-verb-object (SVO) ordering.  We also present results on two distant language pairs, English-Hindi (en-hi) and English-Japanese (ja-en), that follow a SOV word order which is different from the SVO word order of English.
Data statistics are shown in Table~\ref{tab:data} and details are in Appendix~\ref{appendix:dataset}.

\noindent\textbf{Evaluation Method:}
For evaluating alignment performance, it is necessary that the target sentence is exactly the same as for which the gold alignments are provided. Thus, for the alignment experiments,  we force the output token to be from the gold target and only infer the alignment. We then report the Alignment Error Rate (AER)~\citep{och-ney-2000-improved} between the gold alignments and the predicted alignments for different methods. Though our focus is online alignment, for comparison to previous works, we also report results on bidirectional symmetrized alignments in Appendix~\ref{appendix:bidir}.

\noindent\textbf{Methods compared}: We compare our method with both existing statistical alignment models, namely GIZA++~\citep{och-ney-2003-systematic} and FastAlign~\citep{dyer-etal-2013-simple}, and recent Transformer-based alignment methods of \citet{garg-etal-2019-jointly} (\naiveatt) and \citet{chen-etal-2020-accurate} (\shiftatt\ and \shiftaet).
\citet{chen-etal-2020-accurate} also propose a variant of \shiftatt\ called \shiftaet\ that delays computations by one time-step as in \shiftatt, and additionally includes a learned attention sub-layer to compute alignment probabilities.
We also present results on \prioratt\ which is similar to \postatt\ but does not use $\mb y_t$. 
\begin{figure*}[t!]
    \centering
    \includegraphics[width=\textwidth]{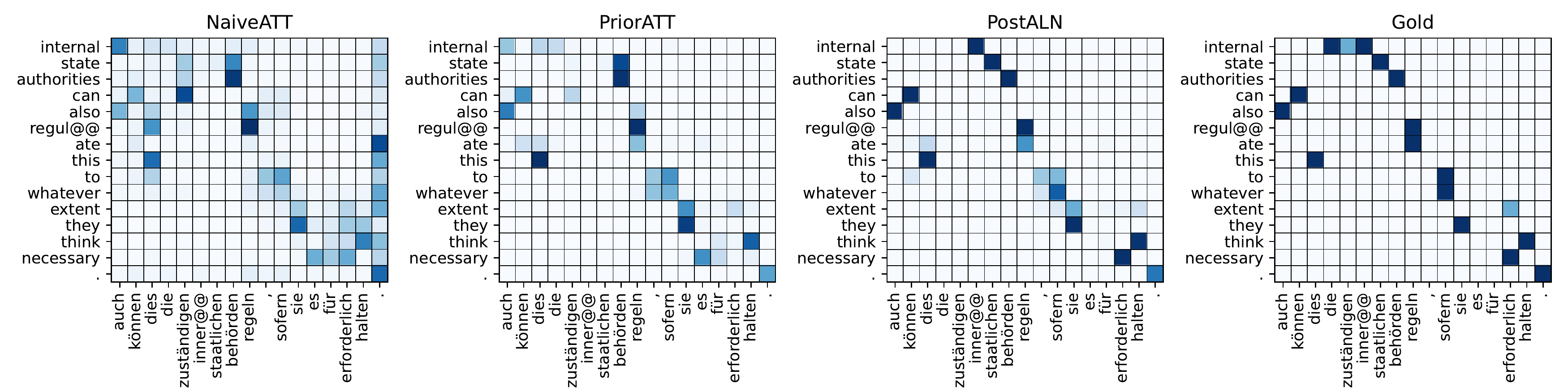}
    \includegraphics[width=\textwidth]{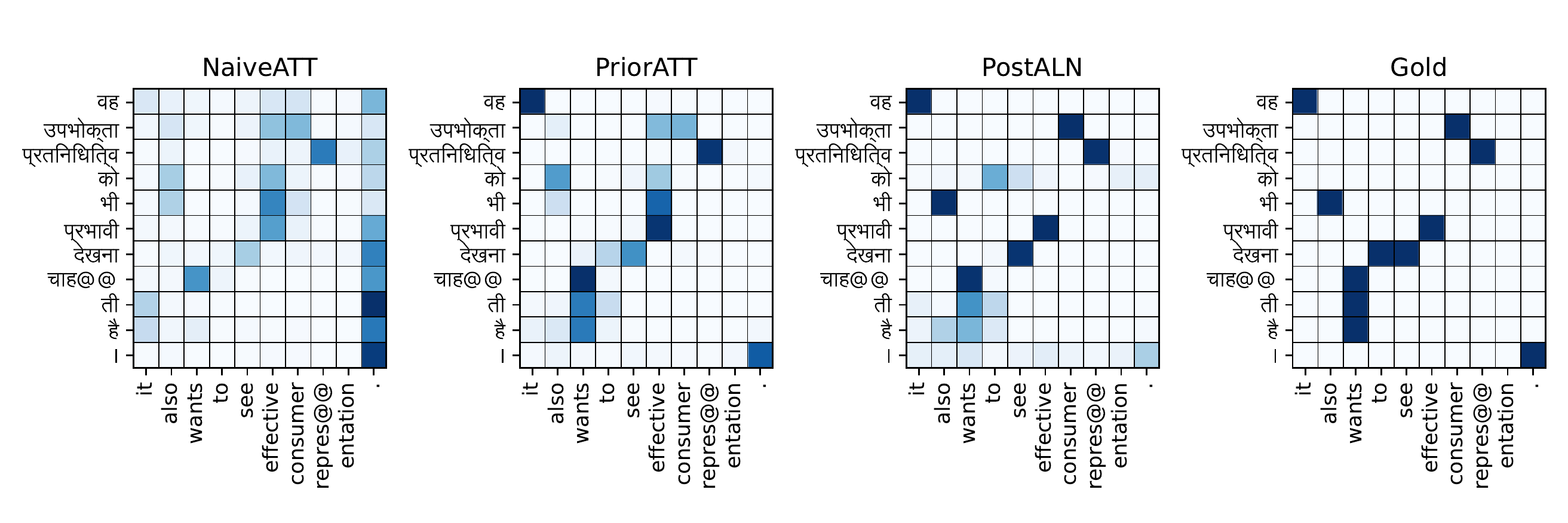}
    \caption{Alignments for de$\rightarrow$en (top-row) and en$\rightarrow$hi (bottom-row) by \naiveatt, \prioratt, and \postatt. Note that \postatt\ is most similar to Gold alignments in the last column.}
    \label{fig:post-vs-prior}
\end{figure*}

\noindent\textbf{Results:}
The alignment results are shown in Table~\ref{tab:aer-main}. First, AERs using statistical methods FastAlign and GIZA++ are shown. Here, for fair comparison, the IBM models used by GIZA++ are trained on the same sub-word units as the Transformer models and sub-word alignments are converted to word level alignments for AER calculations. (GIZA++ has remained a state-of-the-art alignment technique and continues to be compared against.) %
Next, we present alignment results for two vanilla Transformer models - \naiveatt\ and \shiftatt\ - that do not train a separate alignment module. The high AER of \naiveatt\ shows that attention-as-is is very distant from alignment but posterior attention is closer to alignments than prior.  Next we look at methods that train alignment-specific parameters: \prioratt, a prior attention method; \shiftaet\ and \postatt, both posterior alignment methods.  We observe that with training even \prioratt\ has surpassed non-trained posterior. The posterior attention methods outperform the prior attention methods by a large margin, with an improvement of 4.0 to 8.0 points. Within each group, the methods with a trained alignment module outperform the ones without by a huge margin. \postatt\ performs better or matches the performance of \shiftaet\ (achieving the lowest AER in nine out of ten cases in Table~\ref{tab:aer-main}) while avoiding the one-step delay in alignment generation.
Even on the distant languages, \postatt\ achieves significant reductions in error. For ja$\rightarrow$en, we achieve a 1.3 AER reduction compared to \shiftaet\ which is not a truly online method. Figure~\ref{fig:post-vs-prior} shows examples to illustrate the superior alignments of \postatt\ compared to \naiveatt\ and \prioratt.

\begin{table*}
\centering
\tabcolsep 2pt
\resizebox{\textwidth}{!}{
    \begin{tabular}{l|rrrr|rrrr|rrrr|rrrr|rrrr}
           & \multicolumn{4}{c|}{de$\rightarrow$en} & \multicolumn{4}{c|}{en$\rightarrow$fr} & \multicolumn{4}{c|}{ro$\rightarrow$en} & \multicolumn{4}{c|}{en$\rightarrow$hi} & \multicolumn{4}{c}{ja$\rightarrow$en} \\ \cline{2-21} 
    Method & BLEU-C & CSR & BLEU & Time & BLEU-C & CSR & BLEU & Time & BLEU-C & CSR & BLEU & Time & BLEU-C & CSR & BLEU & Time & BLEU-C & CSR & BLEU & Time \\
    \hline \hline
No constraints &  0.0  &  4.6  &  32.9 &  87  &  0.0  &  8.7  &  34.8 &  64  &  0.0  &  8.8  &  33.4 &  47  &   0.0 &   6.3 &  19.7 &  21  &  0.0  &  8.8  &  18.9 &  237 \\ \hline
\naiveatt      &  28.7 &  86.1 &  36.6 &  147 &  36.5 &  88.0 &  38.3 &  93  &  33.3 &  92.3 &  36.5 &  99  &  22.5 &  88.4 &  23.6 &  27  &  15.1 &  75.9 &  20.2 &  315 \\
\prioratt      &  35.0 &  92.8 &  37.6 &  159 &  42.1 &  94.4 &  38.9 &  97  &  36.0 &  91.2 &  37.2 &  100 &  27.2 &  91.5 &  24.4 &  28  &  16.7 &  79.7 &  20.4 &  326 \\ \hline
\shiftatt      &  41.0 &  96.6 &  38.7 &  443 &  45.0 &  93.5 &  38.7 &  239 &  39.2 &  94.2 &  37.4 &  241 &  23.2 &  78.7 &  21.9 &  58  &  15.2 &  72.7 &  19.3 &  567 \\
\shiftaet      &  43.1 &  97.5 &  \h{39.1} &  458 &  46.6 &  94.3 &  39.0 &  235 &  40.8 &  94.4 &  37.6 &  263 &  24.3 &  80.2 &  22.0 &  62  &  18.1 &  75.9 &  19.7 &  596 \\
\postatt       &  42.7 &  97.2 &  39.0 &  399 &  46.3 &  94.1 &  38.7 &  218 &  40.0 &  93.5 &  37.4 &  226 &  23.8 &  79.0 &  22.0 &  47  &  18.2 &  75.7 &  19.7 &  460 \\ \hline
VDBA           &\h{44.5}& 98.9 &  38.5 &  293 &  51.9 &  98.5 &  39.5 &  160 &  43.1 &  99.1 &  37.9 &  165 &  29.8 &  92.3 &  24.5 &  49  &  24.3 &  95.6 &  21.6 &  494 \\
Align-VDBA     &\h{44.5}& 98.6 &  38.6 &  357 &\h{52.9}& 98.4 &  \h{39.7} &  189 &\h{44.1}& 98.9 &  \h{38.1} &  203 &\h{30.5}& 91.5 &  \h{24.7} &  70  &\h{25.1}& 95.5 &  \h{21.8} &  630 \\

\end{tabular}
}
\caption{Constrained translation results showing BLEU-C, CSR (Constraint Satisfaction Rate), BLEU scores and total decoding time (in seconds) for the test set. \alignvdba\ has the highest BLEU-C on all datasets.}
\label{tab:cd}
\end{table*}

\subsection{Impact of \postatt\ on Lexicon-Constrained Translation}
\label{expt:translation}
We next depict the impact of improved AERs from our posterior alignment method on a downstream lexicon-constrained translation task. 
Following previous work~\citep{hokamp-liu-2017-lexically, post-vilar-2018-fast, song-etal-2020-alignment, chen-etal-2020-accurate, chen-etal-2021-lexically}, we extract constraints using the gold alignments and gold translations. Up to three constraints of up to three words each are used for each sentence. Spans correctly translated by a greedy decoding are not selected as constraints.

\noindent
\textbf{Metrics:} Following prior work~\citep{song-etal-2020-alignment}, we report BLEU \citep{papineni-etal-2002-bleu}, time to translate all test sentences, and Constraint Satisfaction Rate (CSR).
However, since it is trivial to get 100\% CSR by always copying, we report another metric to 
evaluate the appropriateness of constraint placement: We call this measure BLEU-C and compute it as the BLEU of the constraint (when satisfied) and a window of three words around it. 
All numbers are averages over five different sets of randomly sampled constraint sets.
The beam size is set to ten by default; results for other beam sizes appear in Appendix~\ref{appendix:more-cd-res}. 

\noindent
\textbf{Methods Compared:} First we compare all the alignment methods presented in Section~\ref{expt:align} on the constrained translation task using the alignment based token-replacement algorithm of \citet{song-etal-2020-alignment} described in Section~\ref{sec:translation-background}. Next, we present a comparison between VBDA~\citep{hu-etal-2019-improved} and our modification \alignvdba.

\begin{table*}[t!]
    \centering
    \resizebox{\textwidth}{!}{
    \begin{tabular}{|l|l|}
    \hline

    Constraints    & (gesetz zur, \textbf{law also}), (dealer, \textbf{pusher}) \\
    Gold           & of course, if a drug addict becomes a \textbf{pusher}, then it is right and necessary that he should pay and answer before the \textbf{law also}. \\
    VDBA           & certainly, if a drug addict becomes a \underline{dealer}, it is right and necessary that he should be brought to justice before the \textbf{law also} \textbf{pusher}. \\
    \alignvdba      & certainly, if a drug addict becomes a \textbf{pusher}, then it is right and necessary that he should be brought to justice before the \textbf{law also}. \\
    
    \hline

    Constraints    & (von mehrheitsverfahren, \textbf{of qualified}) \\
    Gold           & ... whether this is done on the basis of a vote or of consensus, and whether unanimity is required or some form \textbf{of qualified} majority. \\
    VDBA           & ... whether this is done by means \textbf{of qualified} votes or consensus, and whether unanimity or form of majority procedure apply. \\
    \alignvdba       & ... whether this is done by voting or consensus, and whether unanimity or form \textbf{of qualified} majority voting are valid. \\

    \hline

    Constraints    & (zustimmung der, \textbf{strong backing of}) \\
    Gold           & ... which were adopted with the \textbf{strong backing of} the ppe group and the support of the socialist members. \\
    VDBA           & ... which were then adopted with broad agreement from the ppe group and with the \textbf{strong backing of} the socialist members. \\
    \alignvdba     & ... which were then adopted with \textbf{strong backing of} the ppe group and with the support of the socialist members. \\

    \hline
    
    Constraints    & (den usa, \textbf{the usa}), (sicherheitssystems an, \textbf{security system that}), (entwicklung, \textbf{development}) \\
    Gold           & matters we regard as particularly important are improving the working conditions between the weu and the eu \\
                   & and the \textbf{development} of a european \textbf{security system that} is not dependent on \textbf{the usa} . \\
    VDBA           & we consider \textbf{the usa} 's european security system to be particularly important in improving working conditions \\
                   & between the weu and the eu and developing a european \textbf{security system that} is independent of the united states \textbf{development} . \\
    \alignvdba       & we consider the \textbf{development} of the \textbf{security system that} is independent of \textbf{the usa} to be particularly important \\
                   & in improving working conditions between the weu and the eu . \\
    
    \hline

    \end{tabular}
    }
    \caption{Anecdotes showing constrained translations produced by VDBA vs. \alignvdba.}
    \label{tab:anec}
\end{table*}

\noindent\textbf{Results:} Table~\ref{tab:cd} shows that VDBA and our \alignvdba\ that pro-actively enforce constraints have a much higher CSR and BLEU-C compared to the other lazy constraint enforcement methods. For example, for ja$\rightarrow$en greedy methods can only achieve a CSR of 76\% compared to 96\% of the VDBA-based methods. In terms of overall BLEU too, these methods provide an average increase in BLEU of 1.2 and an average increase in BLEU-C of 5 points.  
On average, \alignvdba\ has a 0.7 point greater BLEU-C compared to VDBA. It also has a greater BLEU than VDBA on all the five datasets. In Table~\ref{tab:cd-beam-size} of Appendix we show that for smaller beam-size of 5, the gap between \alignvdba\ and VDBA is even larger (2.1 points greater BLEU-C and 0.4 points greater BLEU).
Table~\ref{tab:anec} lists some example translations by VDBA vs.~\alignvdba. We observe that VDBA places constraints at the end of the translated sentence (e.g., ``pusher", ``development") unlike \alignvdba. In some cases where constraints contain frequent words (like of, the, etc.), VDBA picks the token in the wrong position to tack on the constraint (e.g., ``strong backing of", ``of qualified") while \alignvdba\ places the constraint correctly.

\begin{table}[H]
    \centering
    \resizebox{\linewidth}{!}{
    \begin{tabular}{l|lr|lr}
    Dataset $\rightarrow$ & \multicolumn{2}{c|}{IATE.414} & \multicolumn{2}{c}{Wiktionary.727} \\
    \hline
    Method (Beam Size) $\downarrow$  
                        & BLEU ($\Delta$) & CSR    & BLEU ($\Delta$) & CSR   \\
    \hline\hline
    Baseline (5)        & 25.8          &   76.3 & 26.0            &   76.9 \\
    Train-by-app. (5)   & 26.0 (+0.2)   &   92.9 & 26.9 (+0.9)     &   90.7 \\
    Train-by-rep. (5)   & 26.0 (+0.2)   &   94.5 & 26.3 (+0.3)     &   93.4 \\
    \hline
    No constraints (10) & 29.7          &   77.0 & 29.9            &   72.4 \\
    \shiftaet\ (10)      & 29.9          &   95.9 & 30.4            &   97.2 \\
    VDBA (10)           &\h{30.9}       &   99.8 & 30.9            &   99.4 \\
    Align-VDBA (10)     & \h{30.9 (+1.2)}   &   99.8 &\h{31.1 (+1.2)}  &   99.5 \\ %
    \end{tabular}
    }
    \caption{Constrained translation results on the two real world constraints from \citet{dinu-etal-2019-training}.}
    \label{tab:dinu}
\end{table}
\noindent \textbf{Real World Constraints:}
We also evaluate our method using real world constraints extracted from IATE and Wiktionary datasets by \citet{dinu-etal-2019-training}. 
Table~\ref{tab:dinu} compares \alignvdba\ with the soft-constraints method of \citet{dinu-etal-2019-training} that requires special retraining to teach the model to copy constraints. We reproduced the numbers from their paper in the first three rows. Their baseline is almost 4 BLEU points worse than ours since they used a smaller transformer NMT model, thus making running times incomparable.
When we compare the increment $\Delta$ in BLEU over the respective baselines, \alignvdba\ shows much greater gains of +1.2 vs.~their +0.5. Also, \alignvdba\ provides a larger CSR of 99.6 compared to their 92.
Results for other beam sizes and other methods and metrics appear in Appendix~\ref{appendix:dinu}.

\section{Related Work}
\textbf{Online Prior Alignment from NMTs}: \citet{zenkel-etal-2019-adding} find alignments using a single-head attention submodule, optimized to predict the next token. \citet{garg-etal-2019-jointly} and \citet{song-etal-2020-alignment} supervise a single alignment head from the penultimate multi-head attention with prior alignments from GIZA++ alignments or FastAlign.  \citet{bahar-etal-2020-investigation} and \citet{shankar-etal-2018-surprisingly} treat alignment as a latent variable  and impose a joint distribution over token and alignment while supervising on the token marginal of the joint distribution.

\noindent
\textbf{Online Posterior Alignment from NMTs}: \citet{shankar2018posterior} first identify the role of posterior attention for more accurate alignment.  However, their NMT was a single-headed RNN. \citet{chen-etal-2020-accurate} implement posterior attention in a multi-headed Transformer but they incur a delay of one step between token output and alignment.  We are not aware of any prior work that extracts truly online posterior alignment in modern NMTs.

\noindent
\textbf{Offline Alignment Systems}: Several recent methods apply only in the offline setting: \citet{zenkel-etal-2020-end} extend an NMT with an alignment module; \citet{nagata-etal-2020-supervised} frame alignment  as a question answering task; and \citet{jalili-sabet-etal-2020-simalign,dou-neubig-2021-word} leverage similarity between contextual embeddings from pretrained multilingual models \citep{devlin-etal-2019-bert}.

\noindent
\textbf{Lexicon Constrained Translation}: \citet{hokamp-liu-2017-lexically} and \citet{post-vilar-2018-fast, hu-etal-2019-improved} modify beam search to ensure that target phrases from a given constrained lexicon are present in the translation.  These methods ignore alignment with the source but ensure high success rate for appearance of the target phrases in the constraint. \citet{song-etal-2020-alignment} and \citet{chen-etal-2021-lexically} do consider source alignment but they do not enforce constraints leading to lower CSR. \citet{dinu-etal-2019-training} and \citet{lee-etal-2021-improving} propose alternative training strategies for constraints, whereas we focus on working with existing models. Recently, non autoregressive methods have been proposed for enforcing target constraints but they require that the constraints are given in the order they appear in the target translation~\citep{susanto-etal-2020-lexically}.
\section{Conclusion}
In this paper we proposed a simple architectural modification to modern NMT systems to obtain accurate online alignments. The key idea that led  to high alignment accuracy was conditioning on the output token. Further, our designed alignment module  enables such conditioning to be performed synchronously with token generation. This property led us to \alignvdba, a principled decoding algorithm for lexically constrained translation based on joint distribution of target token and source alignments.  %
Future work includes increase efficiency of constrained inference and harnessing such joint distributions for other forms of constraints, for example, nested constraints.

\noindent{\bf Limitations:} All existing methods for hard constrained inference, including ours, come with considerable runtime overheads. Soft constrained methods are not accurate enough.

\section*{Acknowledgements}
We are grateful to the reviewers for their detailed analysis, thoughtful comments and insightful questions which have helped us improve the paper. We are grateful to Priyesh Jain for providing alignment annotations for 50 English-Hindi sentences.

\bibliography{refs,anthology}

\begin{thebibliography}{46}
\expandafter\ifx\csname natexlab\endcsname\relax\def\natexlab#1{#1}\fi

\bibitem[{Alkhouli et~al.(2018)Alkhouli, Bretschner, and
  Ney}]{alkhouli-etal-2018-alignment}
Tamer Alkhouli, Gabriel Bretschner, and Hermann Ney. 2018.
\newblock \href {https://doi.org/10.18653/v1/W18-6318} {On the alignment
  problem in multi-head attention-based neural machine translation}.
\newblock In \emph{Proceedings of the Third Conference on Machine Translation:
  Research Papers}, pages 177--185, Brussels, Belgium. Association for
  Computational Linguistics.

\bibitem[{Bahar et~al.(2020)Bahar, Makarov, and
  Ney}]{bahar-etal-2020-investigation}
Parnia Bahar, Nikita Makarov, and Hermann Ney. 2020.
\newblock \href {https://aclanthology.org/2020.amta-research.2} {Investigation
  of transformer-based latent attention models for neural machine translation}.
\newblock In \emph{Proceedings of the 14th Conference of the Association for
  Machine Translation in the Americas (Volume 1: Research Track)}, pages 7--20,
  Virtual. Association for Machine Translation in the Americas.

\bibitem[{Bahdanau et~al.(2015)Bahdanau, Cho, and
  Bengio}]{bahdanau-etal-2015-neural}
Dzmitry Bahdanau, Kyunghyun Cho, and Yoshua Bengio. 2015.
\newblock \href {http://arxiv.org/abs/1409.0473} {Neural machine translation by
  jointly learning to align and translate}.
\newblock In \emph{3rd International Conference on Learning Representations,
  {ICLR} 2015, San Diego, CA, USA, May 7-9, 2015, Conference Track
  Proceedings}.

\bibitem[{Brown et~al.(1993)Brown, Della~Pietra, Della~Pietra, and
  Mercer}]{brown-etal-1993-mathematics}
Peter~F. Brown, Stephen~A. Della~Pietra, Vincent~J. Della~Pietra, and Robert~L.
  Mercer. 1993.
\newblock \href {https://aclanthology.org/J93-2003} {The mathematics of
  statistical machine translation: Parameter estimation}.
\newblock \emph{Computational Linguistics}, 19(2):263--311.

\bibitem[{Chen et~al.(2021)Chen, Chen, and Li}]{chen-etal-2021-lexically}
Guanhua Chen, Yun Chen, and Victor~O.K. Li. 2021.
\newblock \href {https://ojs.aaai.org/index.php/AAAI/article/view/17496}
  {Lexically constrained neural machine translation with explicit alignment
  guidance}.
\newblock \emph{Proceedings of the AAAI Conference on Artificial Intelligence},
  35(14):12630--12638.

\bibitem[{Chen et~al.(2020)Chen, Liu, Chen, Jiang, and
  Liu}]{chen-etal-2020-accurate}
Yun Chen, Yang Liu, Guanhua Chen, Xin Jiang, and Qun Liu. 2020.
\newblock \href {https://doi.org/10.18653/v1/2020.emnlp-main.42} {Accurate word
  alignment induction from neural machine translation}.
\newblock In \emph{Proceedings of the 2020 Conference on Empirical Methods in
  Natural Language Processing (EMNLP)}, pages 566--576, Online. Association for
  Computational Linguistics.

\bibitem[{Cho et~al.(2014)Cho, van Merri{\"e}nboer, Bahdanau, and
  Bengio}]{cho-etal-2014-properties}
Kyunghyun Cho, Bart van Merri{\"e}nboer, Dzmitry Bahdanau, and Yoshua Bengio.
  2014.
\newblock \href {https://doi.org/10.3115/v1/W14-4012} {On the properties of
  neural machine translation: Encoder{--}decoder approaches}.
\newblock In \emph{Proceedings of {SSST}-8, Eighth Workshop on Syntax,
  Semantics and Structure in Statistical Translation}, pages 103--111, Doha,
  Qatar. Association for Computational Linguistics.

\bibitem[{Crego et~al.(2016)Crego, Kim, Klein, Rebollo, Yang, Senellart,
  Akhanov, Brunelle, Coquard, Deng, Enoue, Geiss, Johanson, Khalsa, Khiari, Ko,
  Kobus, Lorieux, Martins, Nguyen, Priori, Riccardi, Segal, Servan, Tiquet,
  Wang, Yang, Zhang, Zhou, and Zoldan}]{crego-etal-2016-systrans}
Josep Crego, Jungi Kim, Guillaume Klein, Anabel Rebollo, Kathy Yang, Jean
  Senellart, Egor Akhanov, Patrice Brunelle, Aurelien Coquard, Yongchao Deng,
  Satoshi Enoue, Chiyo Geiss, Joshua Johanson, Ardas Khalsa, Raoum Khiari,
  Byeongil Ko, Catherine Kobus, Jean Lorieux, Leidiana Martins, Dang-Chuan
  Nguyen, Alexandra Priori, Thomas Riccardi, Natalia Segal, Christophe Servan,
  Cyril Tiquet, Bo~Wang, Jin Yang, Dakun Zhang, Jing Zhou, and Peter Zoldan.
  2016.
\newblock \href {http://arxiv.org/abs/1610.05540} {Systran's pure neural
  machine translation systems}.

\bibitem[{Devlin et~al.(2019)Devlin, Chang, Lee, and
  Toutanova}]{devlin-etal-2019-bert}
Jacob Devlin, Ming-Wei Chang, Kenton Lee, and Kristina Toutanova. 2019.
\newblock \href {https://doi.org/10.18653/v1/N19-1423} {{BERT}: Pre-training of
  deep bidirectional transformers for language understanding}.
\newblock In \emph{Proceedings of the 2019 Conference of the North {A}merican
  Chapter of the Association for Computational Linguistics: Human Language
  Technologies, Volume 1 (Long and Short Papers)}, pages 4171--4186,
  Minneapolis, Minnesota. Association for Computational Linguistics.

\bibitem[{Ding et~al.(2019)Ding, Xu, and Koehn}]{ding-etal-2019-saliency}
Shuoyang Ding, Hainan Xu, and Philipp Koehn. 2019.
\newblock \href {https://doi.org/10.18653/v1/W19-5201} {Saliency-driven word
  alignment interpretation for neural machine translation}.
\newblock In \emph{Proceedings of the Fourth Conference on Machine Translation
  (Volume 1: Research Papers)}, pages 1--12, Florence, Italy. Association for
  Computational Linguistics.

\bibitem[{Dinu et~al.(2019)Dinu, Mathur, Federico, and
  Al-Onaizan}]{dinu-etal-2019-training}
Georgiana Dinu, Prashant Mathur, Marcello Federico, and Yaser Al-Onaizan. 2019.
\newblock \href {https://doi.org/10.18653/v1/P19-1294} {Training neural machine
  translation to apply terminology constraints}.
\newblock In \emph{Proceedings of the 57th Annual Meeting of the Association
  for Computational Linguistics}, pages 3063--3068, Florence, Italy.
  Association for Computational Linguistics.

\bibitem[{Dou and Neubig(2021)}]{dou-neubig-2021-word}
Zi-Yi Dou and Graham Neubig. 2021.
\newblock \href {https://aclanthology.org/2021.eacl-main.181} {Word alignment
  by fine-tuning embeddings on parallel corpora}.
\newblock In \emph{Proceedings of the 16th Conference of the European Chapter
  of the Association for Computational Linguistics: Main Volume}, pages
  2112--2128, Online. Association for Computational Linguistics.

\bibitem[{Dyer et~al.(2013)Dyer, Chahuneau, and Smith}]{dyer-etal-2013-simple}
Chris Dyer, Victor Chahuneau, and Noah~A. Smith. 2013.
\newblock \href {https://aclanthology.org/N13-1073} {A simple, fast, and
  effective reparameterization of {IBM} model 2}.
\newblock In \emph{Proceedings of the 2013 Conference of the North {A}merican
  Chapter of the Association for Computational Linguistics: Human Language
  Technologies}, pages 644--648, Atlanta, Georgia. Association for
  Computational Linguistics.

\bibitem[{Garg et~al.(2019)Garg, Peitz, Nallasamy, and
  Paulik}]{garg-etal-2019-jointly}
Sarthak Garg, Stephan Peitz, Udhyakumar Nallasamy, and Matthias Paulik. 2019.
\newblock \href {https://doi.org/10.18653/v1/D19-1453} {Jointly learning to
  align and translate with transformer models}.
\newblock In \emph{Proceedings of the 2019 Conference on Empirical Methods in
  Natural Language Processing and the 9th International Joint Conference on
  Natural Language Processing (EMNLP-IJCNLP)}, pages 4453--4462, Hong Kong,
  China. Association for Computational Linguistics.

\bibitem[{Hasler et~al.(2018)Hasler, de~Gispert, Iglesias, and
  Byrne}]{hasler-etal-2018-neural}
Eva Hasler, Adri{\`a} de~Gispert, Gonzalo Iglesias, and Bill Byrne. 2018.
\newblock \href {https://doi.org/10.18653/v1/N18-2081} {Neural machine
  translation decoding with terminology constraints}.
\newblock In \emph{Proceedings of the 2018 Conference of the North {A}merican
  Chapter of the Association for Computational Linguistics: Human Language
  Technologies, Volume 2 (Short Papers)}, pages 506--512, New Orleans,
  Louisiana. Association for Computational Linguistics.

\bibitem[{Hieber et~al.(2020)Hieber, Domhan, Denkowski, and
  Vilar}]{hieber-etal-2020-sockeye}
Felix Hieber, Tobias Domhan, Michael Denkowski, and David Vilar. 2020.
\newblock \href {https://aclanthology.org/2020.eamt-1.50} {Sockeye 2: A toolkit
  for neural machine translation}.
\newblock In \emph{Proceedings of the 22nd Annual Conference of the European
  Association for Machine Translation}, pages 457--458, Lisboa, Portugal.
  European Association for Machine Translation.

\bibitem[{Hokamp and Liu(2017)}]{hokamp-liu-2017-lexically}
Chris Hokamp and Qun Liu. 2017.
\newblock \href {https://doi.org/10.18653/v1/P17-1141} {Lexically constrained
  decoding for sequence generation using grid beam search}.
\newblock In \emph{Proceedings of the 55th Annual Meeting of the Association
  for Computational Linguistics (Volume 1: Long Papers)}, pages 1535--1546,
  Vancouver, Canada. Association for Computational Linguistics.

\bibitem[{Hu et~al.(2019)Hu, Khayrallah, Culkin, Xia, Chen, Post, and
  Van~Durme}]{hu-etal-2019-improved}
J.~Edward Hu, Huda Khayrallah, Ryan Culkin, Patrick Xia, Tongfei Chen, Matt
  Post, and Benjamin Van~Durme. 2019.
\newblock \href {https://doi.org/10.18653/v1/N19-1090} {Improved lexically
  constrained decoding for translation and monolingual rewriting}.
\newblock In \emph{Proceedings of the 2019 Conference of the North {A}merican
  Chapter of the Association for Computational Linguistics: Human Language
  Technologies, Volume 1 (Long and Short Papers)}, pages 839--850, Minneapolis,
  Minnesota. Association for Computational Linguistics.

\bibitem[{Jalili~Sabet et~al.(2020)Jalili~Sabet, Dufter, Yvon, and
  Sch{\"u}tze}]{jalili-sabet-etal-2020-simalign}
Masoud Jalili~Sabet, Philipp Dufter, Fran{\c{c}}ois Yvon, and Hinrich
  Sch{\"u}tze. 2020.
\newblock \href {https://doi.org/10.18653/v1/2020.findings-emnlp.147}
  {{S}im{A}lign: High quality word alignments without parallel training data
  using static and contextualized embeddings}.
\newblock In \emph{Findings of the Association for Computational Linguistics:
  EMNLP 2020}, pages 1627--1643, Online. Association for Computational
  Linguistics.

\bibitem[{Kalchbrenner and Blunsom(2013)}]{kalchbrenner-blunsom-2013-recurrent}
Nal Kalchbrenner and Phil Blunsom. 2013.
\newblock \href {https://aclanthology.org/D13-1176} {Recurrent continuous
  translation models}.
\newblock In \emph{Proceedings of the 2013 Conference on Empirical Methods in
  Natural Language Processing}, pages 1700--1709, Seattle, Washington, USA.
  Association for Computational Linguistics.

\bibitem[{Koehn(2004)}]{koehn-2004-statistical}
Philipp Koehn. 2004.
\newblock \href {https://aclanthology.org/W04-3250} {Statistical significance
  tests for machine translation evaluation}.
\newblock In \emph{Proceedings of the 2004 Conference on Empirical Methods in
  Natural Language Processing}, pages 388--395, Barcelona, Spain. Association
  for Computational Linguistics.

\bibitem[{Koehn et~al.(2005)Koehn, Axelrod, Mayne, Callison-Burch, Osborne, and
  Talbot}]{koehn-etal-2005-edinburgh}
Philipp Koehn, Amittai Axelrod, Alexandra~Birch Mayne, Chris Callison-Burch,
  Miles Osborne, and David Talbot. 2005.
\newblock \href
  {https://www.isca-speech.org/archive/iwslt_05/papers/slt5_068.pdf} {Edinburgh
  system description for the 2005 iwslt speech translation evaluation}.
\newblock In \emph{International Workshop on Spoken Language Translation
  (IWSLT) 2005}.

\bibitem[{Kunchukuttan et~al.(2018)Kunchukuttan, Mehta, and
  Bhattacharyya}]{kunchukuttan-etal-2018-iit}
Anoop Kunchukuttan, Pratik Mehta, and Pushpak Bhattacharyya. 2018.
\newblock \href {https://aclanthology.org/L18-1548} {The {IIT} {B}ombay
  {E}nglish-{H}indi parallel corpus}.
\newblock In \emph{Proceedings of the Eleventh International Conference on
  Language Resources and Evaluation ({LREC} 2018)}, Miyazaki, Japan. European
  Language Resources Association (ELRA).

\bibitem[{Lee et~al.(2021)Lee, Yang, and Choi}]{lee-etal-2021-improving}
Gyubok Lee, Seongjun Yang, and Edward Choi. 2021.
\newblock \href {https://doi.org/10.18653/v1/2021.acl-short.94} {Improving
  lexically constrained neural machine translation with source-conditioned
  masked span prediction}.
\newblock In \emph{Proceedings of the 59th Annual Meeting of the Association
  for Computational Linguistics and the 11th International Joint Conference on
  Natural Language Processing (Volume 2: Short Papers)}, pages 743--753,
  Online. Association for Computational Linguistics.

\bibitem[{Martin et~al.(2005)Martin, Mihalcea, and
  Pedersen}]{martin-etal-2005-word}
Joel Martin, Rada Mihalcea, and Ted Pedersen. 2005.
\newblock \href {https://aclanthology.org/W05-0809} {Word alignment for
  languages with scarce resources}.
\newblock In \emph{Proceedings of the {ACL} Workshop on Building and Using
  Parallel Texts}, pages 65--74, Ann Arbor, Michigan. Association for
  Computational Linguistics.

\bibitem[{Mihalcea and Pedersen(2003)}]{mihalcea-pedersen-2003-evaluation}
Rada Mihalcea and Ted Pedersen. 2003.
\newblock \href {https://aclanthology.org/W03-0301} {An evaluation exercise for
  word alignment}.
\newblock In \emph{Proceedings of the {HLT}-{NAACL} 2003 Workshop on Building
  and Using Parallel Texts: Data Driven Machine Translation and Beyond}, pages
  1--10.

\bibitem[{M{\"u}ller(2017)}]{muller-2017-treatment}
Mathias M{\"u}ller. 2017.
\newblock \href {https://doi.org/10.18653/v1/W17-4804} {Treatment of markup in
  statistical machine translation}.
\newblock In \emph{Proceedings of the Third Workshop on Discourse in Machine
  Translation}, pages 36--46, Copenhagen, Denmark. Association for
  Computational Linguistics.

\bibitem[{Nagata et~al.(2020)Nagata, Chousa, and
  Nishino}]{nagata-etal-2020-supervised}
Masaaki Nagata, Katsuki Chousa, and Masaaki Nishino. 2020.
\newblock \href {https://doi.org/10.18653/v1/2020.emnlp-main.41} {A supervised
  word alignment method based on cross-language span prediction using
  multilingual {BERT}}.
\newblock In \emph{Proceedings of the 2020 Conference on Empirical Methods in
  Natural Language Processing (EMNLP)}, pages 555--565, Online. Association for
  Computational Linguistics.

\bibitem[{Neubig(2011)}]{neubig-2011-kyoto}
Graham Neubig. 2011.
\newblock \href {http://www.phontron.com/kftt} {The {Kyoto} free translation
  task}.

\bibitem[{Och and Ney(2000)}]{och-ney-2000-improved}
Franz~Josef Och and Hermann Ney. 2000.
\newblock \href {https://doi.org/10.3115/1075218.1075274} {Improved statistical
  alignment models}.
\newblock In \emph{Proceedings of the 38th Annual Meeting of the Association
  for Computational Linguistics}, pages 440--447, Hong Kong. Association for
  Computational Linguistics.

\bibitem[{Och and Ney(2003)}]{och-ney-2003-systematic}
Franz~Josef Och and Hermann Ney. 2003.
\newblock \href {https://doi.org/10.1162/089120103321337421} {A systematic
  comparison of various statistical alignment models}.
\newblock \emph{Computational Linguistics}, 29(1):19--51.

\bibitem[{Ott et~al.(2019)Ott, Edunov, Baevski, Fan, Gross, Ng, Grangier, and
  Auli}]{ott-etal-2019-fairseq}
Myle Ott, Sergey Edunov, Alexei Baevski, Angela Fan, Sam Gross, Nathan Ng,
  David Grangier, and Michael Auli. 2019.
\newblock \href {https://doi.org/10.18653/v1/N19-4009} {fairseq: A fast,
  extensible toolkit for sequence modeling}.
\newblock In \emph{Proceedings of the 2019 Conference of the North {A}merican
  Chapter of the Association for Computational Linguistics (Demonstrations)},
  pages 48--53, Minneapolis, Minnesota. Association for Computational
  Linguistics.

\bibitem[{Papineni et~al.(2002)Papineni, Roukos, Ward, and
  Zhu}]{papineni-etal-2002-bleu}
Kishore Papineni, Salim Roukos, Todd Ward, and Wei-Jing Zhu. 2002.
\newblock \href {https://doi.org/10.3115/1073083.1073135} {{B}leu: a method for
  automatic evaluation of machine translation}.
\newblock In \emph{Proceedings of the 40th Annual Meeting of the Association
  for Computational Linguistics}, pages 311--318, Philadelphia, Pennsylvania,
  USA. Association for Computational Linguistics.

\bibitem[{Post(2018)}]{post-2018-call}
Matt Post. 2018.
\newblock \href {https://doi.org/10.18653/v1/W18-6319} {A call for clarity in
  reporting {BLEU} scores}.
\newblock In \emph{Proceedings of the Third Conference on Machine Translation:
  Research Papers}, pages 186--191, Brussels, Belgium. Association for
  Computational Linguistics.

\bibitem[{Post and Vilar(2018)}]{post-vilar-2018-fast}
Matt Post and David Vilar. 2018.
\newblock \href {https://doi.org/10.18653/v1/N18-1119} {Fast lexically
  constrained decoding with dynamic beam allocation for neural machine
  translation}.
\newblock In \emph{Proceedings of the 2018 Conference of the North {A}merican
  Chapter of the Association for Computational Linguistics: Human Language
  Technologies, Volume 1 (Long Papers)}, pages 1314--1324, New Orleans,
  Louisiana. Association for Computational Linguistics.

\bibitem[{Sennrich et~al.(2016)Sennrich, Haddow, and
  Birch}]{sennrich-etal-2016-neural}
Rico Sennrich, Barry Haddow, and Alexandra Birch. 2016.
\newblock \href {https://doi.org/10.18653/v1/P16-1162} {Neural machine
  translation of rare words with subword units}.
\newblock In \emph{Proceedings of the 54th Annual Meeting of the Association
  for Computational Linguistics (Volume 1: Long Papers)}, pages 1715--1725,
  Berlin, Germany. Association for Computational Linguistics.

\bibitem[{Shankar et~al.(2018)Shankar, Garg, and
  Sarawagi}]{shankar-etal-2018-surprisingly}
Shiv Shankar, Siddhant Garg, and Sunita Sarawagi. 2018.
\newblock \href {https://doi.org/10.18653/v1/D18-1065} {Surprisingly easy
  hard-attention for sequence to sequence learning}.
\newblock In \emph{Proceedings of the 2018 Conference on Empirical Methods in
  Natural Language Processing}, pages 640--645, Brussels, Belgium. Association
  for Computational Linguistics.

\bibitem[{Shankar and Sarawagi(2019)}]{shankar2018posterior}
Shiv Shankar and Sunita Sarawagi. 2019.
\newblock \href {https://openreview.net/forum?id=BkltNhC9FX} {Posterior
  attention models for sequence to sequence learning}.
\newblock In \emph{International Conference on Learning Representations}.

\bibitem[{Shen et~al.(2019)Shen, Zhao, Su, and
  Klakow}]{shen-etal-2019-improving}
Xiaoyu Shen, Yang Zhao, Hui Su, and Dietrich Klakow. 2019.
\newblock \href {https://doi.org/10.18653/v1/D19-1390} {Improving latent
  alignment in text summarization by generalizing the pointer generator}.
\newblock In \emph{Proceedings of the 2019 Conference on Empirical Methods in
  Natural Language Processing and the 9th International Joint Conference on
  Natural Language Processing (EMNLP-IJCNLP)}, pages 3762--3773, Hong Kong,
  China. Association for Computational Linguistics.

\bibitem[{Song et~al.(2020)Song, Wang, Yu, Zhang, Huang, Luo, Duan, and
  Zhang}]{song-etal-2020-alignment}
Kai Song, Kun Wang, Heng Yu, Yue Zhang, Zhongqiang Huang, Weihua Luo, Xiangyu
  Duan, and Min Zhang. 2020.
\newblock \href {https://doi.org/10.1609/aaai.v34i05.6418} {Alignment-enhanced
  transformer for constraining nmt with pre-specified translations}.
\newblock \emph{Proceedings of the AAAI Conference on Artificial Intelligence},
  34(05):8886--8893.

\bibitem[{Susanto et~al.(2020)Susanto, Chollampatt, and
  Tan}]{susanto-etal-2020-lexically}
Raymond~Hendy Susanto, Shamil Chollampatt, and Liling Tan. 2020.
\newblock \href {https://doi.org/10.18653/v1/2020.acl-main.325} {Lexically
  constrained neural machine translation with {L}evenshtein transformer}.
\newblock In \emph{Proceedings of the 58th Annual Meeting of the Association
  for Computational Linguistics}, pages 3536--3543, Online. Association for
  Computational Linguistics.

\bibitem[{Sutskever et~al.(2014)Sutskever, Vinyals, and
  Le}]{sutskever-etal-2014-sequence}
Ilya Sutskever, Oriol Vinyals, and Quoc~V Le. 2014.
\newblock \href
  {https://proceedings.neurips.cc/paper/2014/file/a14ac55a4f27472c5d894ec1c3c743d2-Paper.pdf}
  {Sequence to sequence learning with neural networks}.
\newblock In \emph{Advances in Neural Information Processing Systems},
  volume~27. Curran Associates, Inc.

\bibitem[{Vaswani et~al.(2017)Vaswani, Shazeer, Parmar, Uszkoreit, Jones,
  Gomez, Kaiser, and Polosukhin}]{vaswani-etal-2017-attention}
Ashish Vaswani, Noam Shazeer, Niki Parmar, Jakob Uszkoreit, Llion Jones,
  Aidan~N Gomez, {\L}ukasz Kaiser, and Illia Polosukhin. 2017.
\newblock \href
  {https://proceedings.neurips.cc/paper/2017/file/3f5ee243547dee91fbd053c1c4a845aa-Paper.pdf}
  {Attention is all you need}.
\newblock In \emph{Advances in Neural Information Processing Systems},
  volume~30. Curran Associates, Inc.

\bibitem[{Vilar et~al.(2006)Vilar, Popovi{\'c}, and Ney}]{vilar-etal-2006-aer}
David Vilar, Maja Popovi{\'c}, and Hermann Ney. 2006.
\newblock \href
  {https://www-i6.informatik.rwth-aachen.de/publications/download/277/Vilar-IWSLT-2006.pdf}
  {{AER}: Do we need to “improve” our alignments?}
\newblock In \emph{International Workshop on Spoken Language Translation
  (IWSLT) 2006}.

\bibitem[{Zenkel et~al.(2019)Zenkel, Wuebker, and
  DeNero}]{zenkel-etal-2019-adding}
Thomas Zenkel, Joern Wuebker, and John DeNero. 2019.
\newblock \href {http://arxiv.org/abs/1901.11359} {Adding interpretable
  attention to neural translation models improves word alignment}.

\bibitem[{Zenkel et~al.(2020)Zenkel, Wuebker, and
  DeNero}]{zenkel-etal-2020-end}
Thomas Zenkel, Joern Wuebker, and John DeNero. 2020.
\newblock \href {https://doi.org/10.18653/v1/2020.acl-main.146} {End-to-end
  neural word alignment outperforms {GIZA}++}.
\newblock In \emph{Proceedings of the 58th Annual Meeting of the Association
  for Computational Linguistics}, pages 1605--1617, Online. Association for
  Computational Linguistics.

\end{thebibliography}
\bibliographystyle{acl_natbib}

\clearpage
\appendix

\section{Alignment Error Rate}
\label{appendix:aer}
Given gold alignments consisting of sure alignments $\mcal{S}$ and possible alignments $\mcal{P}$, and the predicted alignments $\calA$, the Alignment Error Rate (AER) is defined as \citep{och-ney-2000-improved}:
\begin{equation*}
    \text{AER} = 1 - \frac{|\calA\cap\mcal{P}|+|\calA\cap\mcal{S}|}{|\calA| + |\mcal{S}|}
\end{equation*}
Note that here $\mcal{S} \subseteq \mcal{P}$. Also note that since our models are trained on sub-word units but gold alignments are over words, we need to convert alignments between word pieces to alignments between words. A source word and a target word are said to be aligned if there exists an alignment link between any of their respective word pieces.

\section{BLEU-C}
\label{appendix:spanbleu}
\begin{table*}[ht]
    \centering
    \resizebox{0.8\textwidth}{!}{
    \begin{tabular}{l|l|l}
         Reference & \multicolumn{2}{l}{we consider the \textbf{development} of a robust \textbf{security system} that is independent of the} \\
         Prediction & \multicolumn{2}{l}{we consider developing a robust \textbf{security system} which is independent of the} \\
         \hline
         \multicolumn{3}{c}{BLEU-C (Window Size = 2)} \\
         \hline
         Cons. No & Reference Spans & Predicted Spans \\
         \hline
         1 & consider the \textbf{development} of a & (empty sentence) \\
         2 & a robust \textbf{security system} that is & a robust \textbf{security system} which is \\
         \hline
         \multicolumn{3}{l}{BLEU-C = BLEU(Reference Spans, Predicted Spans)}
    \end{tabular}
    }
    \caption{An example BLEU-C computation}
    \label{tab:spanbleu}
\end{table*}

Given a reference sentence, a predicted translation and a set of constraints, for each constraints, a segment of the sentence is chosen which contains the constraint and window size words (if available) surrounding the constraint words on either side. Such segments, called spans, are collected for the reference and predicted sentences in the test set and BLEU is computed over these spans. If a constraint is not satisfied in the prediction, the corresponding span is considered to be the empty string. An example is shown in Table~\ref{tab:spanbleu}. Table~\ref{tab:spanBLEUwins} shows how BLEU-C varies as a function of varying window size for a fixed English-French constraint set with beam size set to 10.

\begin{table}[H]
    \centering
    \resizebox{\linewidth}{!}{
    \begin{tabular}{l|rrrrrrr}
Window Size $\rightarrow$ & 2 & 3 & 4 & 5 & 6 & 7 & 8 \\
\hline
No constraints  &  0.0 &  0.0 &  0.0 &  0.0 &  0.0 &  0.0 &  0.0 \\
\naiveatt       & 34.4 & 32.0 & 30.4 & 29.5 & 29.4 & 29.5 & 29.7 \\
\prioratt       & 41.5 & 38.7 & 36.4 & 35.1 & 34.9 & 35.0 & 35.2 \\
\shiftatt       & 44.9 & 41.5 & 38.9 & 37.3 & 36.4 & 36.2 & 36.0 \\
\shiftaet       & 47.0 & 43.2 & 40.4 & 38.7 & 38.0 & 37.6 & 37.4 \\
\postatt        & 46.4 & 42.7 & 39.8 & 38.0 & 37.1 & 36.9 & 36.6 \\
VDBA            & 54.9 & 50.5 & 46.8 & 44.6 & 43.5 & 43.0 & 42.6 \\
Align-VDBA      & 56.4 & 51.7 & 47.9 & 45.6 & 44.4 & 43.7 & 43.3
    \end{tabular}
    }
    \caption{BLEU-C vs Window Size}
    \label{tab:spanBLEUwins}
\end{table}

\section{Description of the Datasets}%
\label{appendix:dataset}
The European languages consist of parallel sentences for three language pairs from the Europarl Corpus and alignments from \citet{mihalcea-pedersen-2003-evaluation}, \citet{och-ney-2000-improved}, \citet{vilar-etal-2006-aer}. 
Following previous works \citep{ding-etal-2019-saliency, chen-etal-2020-accurate}, the last 1000 sentences of the training data are used as validation data. 

For English-Hindi, we use the dataset from \citet{martin-etal-2005-word} consisting of 3440 training sentence pairs, 25 validation and 90 test sentences with gold alignments. Since training Transformers requires much larger datasets, we augment the training set with 1.6 million sentences from the IIT Bombay Parallel Corpus \citep{kunchukuttan-etal-2018-iit}. We also add the first 50 sentences from the dev set of IIT Bombay Parallel Corpus with manually annotated alignments to the test set giving a total of 140 test sentences.

For Japanese-English, we use The Kyoto Free Translation Task~\citep{neubig-2011-kyoto}. It comprises roughly 330K training, 1166 validation and 1235 test sentences. As with other datasets, gold alignments are available only for the test sentences. The Japanese text is already segmented and we use it without additional changes. 

The real world constraints datasets of \citet{dinu-etal-2019-training} are extracted from the German-English WMT newstest 2017 task with the IATE dataset consisting of 414 sentences (451 constraints) and the Wiktionary 727 sentences (879 constraints). The constraints come from the IATE and Wiktionary termninology databases.

All datasets were processed using the scripts provided by \citet{zenkel-etal-2019-adding} at \url{https://github.com/lilt/alignment-scripts}. Computation of BLEU and BLEU-C, and the paired test were performed using \texttt{sacrebleu}~\citep{post-2018-call}.

\section{Bidirectional Symmetrized Alignment}
\label{appendix:bidir}

We report AERs using bidirectional symmetrized alignments in Table~\ref{tab:bidir} in order to provide fair comparisons to results in prior literature. The symmetrization is done using the \textit{grow-diagonal} heuristic \citep{koehn-etal-2005-edinburgh, och-ney-2000-improved}. Since bidirectional alignments need the entire text in both languages, these are not  online alignments.

\begin{table}[H]
\tabcolsep 2pt
\centering
\resizebox{0.8\linewidth}{!}{
    \begin{tabular}{l|rrrrr}
    Method & de-en & en-fr & ro-en & en-hi & ja-en \\ 
    \hline \hline
    \multicolumn{6}{c}{Statistical Methods} \\
    \hline
    GIZA++       & 18.6 &  \h{5.5} & 26.3 & 35.9 & 39.7 \\
    FastAlign    & 27.0 & 10.5 & 32.1 &    - &    - \\
    \hline
    \multicolumn{6}{c}{No Alignment Training} \\
    \hline
    \naiveatt    & 29.2 & 16.9 & 31.4 & 43.8 & 57.1 \\
    \shiftatt    & 16.9 &  7.8 & 24.3 & 30.9 & 46.2 \\
    \hline
    \multicolumn{6}{c}{With Alignment Training} \\
    \hline
    \prioratt    & 22.0 & 10.1 & 26.3 & 32.1 & 48.2 \\
    \shiftaet    & 15.4 &  5.6 & \textbf{21.0} & 26.7 & 40.1\\
    \postatt     & \textbf{15.3} &  \textbf{5.5} & \textbf{21.0} & \textbf{26.1} & \textbf{39.5}\\
    \end{tabular}
}
\caption{AERs for bidirectional symmetrized alignments. \postatt\ consistently performs the best.}
\label{tab:bidir}
\end{table}

\begin{table*}[ht]
\tabcolsep 3pt
    \centering
    \resizebox{\textwidth}{!}{
    \begin{tabular}{l|l|rrrr|rrrr|rrrr|rrrr|rrrr}
 \multicolumn{2}{c|}{} & \multicolumn{4}{c|}{de$\rightarrow$en} & \multicolumn{4}{c|}{en$\rightarrow$fr} & \multicolumn{4}{c|}{ro$\rightarrow$en} & \multicolumn{4}{c|}{en$\rightarrow$hi} & \multicolumn{4}{c}{ja$\rightarrow$en} \\
\cline{3-22} 
\begin{tabular}[c]{@{}l@{}}Beam\\Size\end{tabular} & Method & BLEU-C & CSR & BLEU & Time & BLEU-C & CSR & BLEU & Time & BLEU-C & CSR & BLEU & Time & BLEU-C & CSR & BLEU & Time & BLEU-C & CSR & BLEU & Time \\
\hline
5& No constraints &  0.0  &  5.0  &  32.9 &  78  &  0.0  &  8.7  &  34.6 &  61  &  0.0  &  8.4  &  33.3 &  45  &  0.0  &   5.6 &  19.7 &  18  &  0.0  &  7.9  &  19.1 &  221 \\
&  \naiveatt      &  28.9 &  86.2 &  36.7 &  127 &  36.7 &  88.6 &  38.0 &  87  &  32.9 &  91.8 &  36.3 &  88  &  23.0 &  89.9 &  23.9 &  25  &  15.1 &  77.0 &  20.3 &  398 \\
&  \prioratt      &  35.3 &  93.0 &  37.7 &  136 &  42.2 &  94.7 &  38.6 &  89  &  36.0 &  91.6 &  37.0 &  89  &  27.6 &  91.7 &  {24.7}&  26  &  16.8 &  80.2 &  20.6 &  353 \\
&  \shiftatt      &  41.0 &  96.7 &  38.7 &  268 &  45.2 &  93.8 &  38.4 &  167 &  39.2 &  94.4 &  37.2 &  160 &  23.8 &  81.8 &  22.0 &  42  &  15.1 &  72.6 &  19.3 &  664 \\
&  \shiftaet      &  {43.1}&  97.6 &  {39.1}&  291 &  46.5 &  94.8 &  38.6 &  165 &  40.8 &  94.7 &  {37.5}&  163 &  24.5 &  83.6 &  22.1 &  44  &  18.0 &  76.5 &  19.6 &  583 \\
&  \postatt       &  42.7 &  97.3 &  39.0 &  252 &  46.1 &  93.9 &  38.5 &  151 &  39.8 &  93.5 &  37.3 &  141 &  23.3 &  79.7 &  21.7 &  39  &  17.9 &  75.3 &  19.6 &  469 \\
&  VDBA           &  39.6 &  99.4 &  37.8 &  203 &  45.9 &  99.5 &  38.5 &  109 &  36.6 &  99.2 &  36.7 &  117 &  27.3 &  96.6 &  24.2 &  37  &  22.1 &  96.9 &  20.9 &  397 \\
& \alignvdba*     &  40.3 &  99.0 &  38.0 &  244 &  47.4 &  99.3 &  38.7 &  132 &  37.6 &  99.7 &  36.8 &  139 &  27.2 &  95.6 &  24.1 &  46  &  22.5 &  97.2 &  21.0 &  460 \\
& \alignvdba      &  41.3 &  98.8 &  38.2 &	 236 &  {48.0}&  98.9 &  {38.7}&  128 &  {42.0}&  96.6 &  {37.5}&  134 &  {28.2}&  91.3 &  {24.7}&  45 &  {22.6}&  93.9 &  {21.2}&  445    \\ %
\hline
10&No constraints &  0.0  &  4.6  &  32.9 &  87  &  0.0  &  8.7  &  34.8 &  64  &  0.0  &  8.8  &  33.4 &  47  &   0.0 &   6.3 &  19.7 &  21  &  0.0  &  8.8  &  18.9 &  237 \\
&  \naiveatt      &  28.7 &  86.1 &  36.6 &  147 &  36.5 &  88.0 &  38.3 &  93  &  33.3 &  92.3 &  36.5 &  99  &  22.5 &  88.4 &  23.6 &  27  &  15.1 &  75.9 &  20.2 &  315 \\
&  \prioratt      &  35.0 &  92.8 &  37.6 &  159 &  42.1 &  94.4 &  38.9 &  97  &  36.0 &  91.2 &  37.2 &  100 &  27.2 &  91.5 &  24.4 &  28  &  16.7 &  79.7 &  20.4 &  326 \\
&  \shiftatt      &  41.0 &  96.6 &  38.7 &  443 &  45.0 &  93.5 &  38.7 &  239 &  39.2 &  94.2 &  37.4 &  241 &  23.2 &  78.7 &  21.9 &  58  &  15.2 &  72.7 &  19.3 &  567 \\
&  \shiftaet      &  43.1 &  97.5 &  {39.1}&  458 &  46.6 &  94.3 &  39.0 &  235 &  40.8 &  94.4 &  37.6 &  263 &  24.3 &  80.2 &  22.0 &  62  &  18.1 &  75.9 &  19.7 &  596 \\
&  \postatt       &  42.7 &  97.2 &  39.0 &  399 &  46.3 &  94.1 &  38.7 &  218 &  40.0 &  93.5 &  37.4 &  226 &  23.8 &  79.0 &  22.0 &  47  &  18.2 &  75.7 &  19.7 &  460 \\
&  VDBA           &  {44.5}&  98.9 &  38.5 &  293 &  51.9 &  98.5 &  39.5 &  160 &  43.1 &  99.1 &  37.9 &  165 &  29.8 &  92.3 &  24.5 &  49  &  24.3 &  95.6 &  21.6 &  494 \\
&  Align-VDBA     &  {44.5}&  98.6 &  38.6 &  357 &  {52.9}&  98.4 &  {39.7}&  189 &  {44.1}&  98.9 &  {38.1}&  203 &  {30.5}&  91.5 &  {24.7}&  70  &  {25.1}&  95.5 &  {21.8}&  630 \\
    \end{tabular}
    }
    \caption{Lexically Constrained Translation Results with different beam sizes. All numbers are average over 5 randomly sampled constraint sets and running times are in seconds. \alignvdba* denotes \alignvdba\ without alignment probability based beam allocation (\textit{i.e.} with threshold set to 0).}
    \label{tab:cd-beam-size}
\end{table*}

\begin{table}[h]
    \centering
\resizebox{\linewidth}{!}{
\begin{tabular}{l|l|l|l}
\multicolumn{1}{c|}{} & \multicolumn{1}{c|}{de$\rightarrow$en} & \multicolumn{1}{c|}{en$\rightarrow$fr} & \multicolumn{1}{c}{ro$\rightarrow$en} \\ %
\hline
No constraints&    0.0001* & 0.0001* & 0.0001* \\
\naiveatt     &    0.0001* & 0.0001* & 0.0001* \\
\prioratt     &    0.0001* & 0.0001* & 0.0001* \\
\shiftatt     &     0.1700 & 0.0001* & 0.0001* \\
\shiftaet     &    0.0015* & 0.0001* & 0.0018* \\
\postatt      &    0.0032* & 0.0001* & 0.0003* \\
VDBA          &     0.2666 & 0.0020* & 0.0229* \\
\end{tabular}
}
\caption{$p$-values from paired bootstrap resampling tests with 10000 bootstrap samples for BLEU on Table~\ref{tab:cd} datasets for beam size 10. Tests are performed with respect to \alignvdba. * denotes statistically significant difference from \alignvdba\ at power 0.05 (p-value < 0.05).}
\label{tab:cd-test}
\end{table}

\section{Additional Lexicon-Constrained Translation Results}
\label{appendix:more-cd-res}
Constrained translation results for beam sizes 5 and 10 are shown in Table~\ref{tab:cd-beam-size}. We also present results for \alignvdba\ without the alignment probability based beam allocation as \alignvdba* in Table~\ref{tab:cd-beam-size}. We can see that our beam allocation technique results in better beam utilization as evidenced by improvements in BLEU and BLEU-C, and reduction total decoding time.

Paired bootstrap resampling test~\citep{koehn-2004-statistical} results with respect to \alignvdba\ for beam size 10 are shown in Table~\ref{tab:cd-test}.

\section{Additional Real World Constrained Translation Results}
\label{appendix:dinu}
Results on the real world constrained translation datasets of \citet{dinu-etal-2019-training} for all the methods in Table~\ref{tab:cd} with beam sizes 5, 10 and 20 are presented in Table~\ref{tab:dinu-all}. Paired bootstrap resampling test~\citep{koehn-2004-statistical} results with respect to \alignvdba\ for beam size 5 are shown in Table~\ref{tab:dinu-test}

\begin{table}[h]
\tabcolsep 2pt
    \centering
    \resizebox{\linewidth}{!}{
    \begin{tabular}{l|l|rrrr|rrrr}
     & Dataset $\rightarrow$ & \multicolumn{4}{c|}{IATE.414} & \multicolumn{4}{c}{Wiktionary.727} \\
    \hline
    \begin{tabular}[c]{@{}l@{}}Beam\\Size\end{tabular} & Method $\downarrow$  & BLEU-C & CSR & BLEU & Time& BLEU-C & CSR & BLEU & Time \\
    \hline
    5 & No constraints & 27.9 & 76.6 & 29.7 & 134 & 26.3 & 72.0 & 29.9 & 217 \\
     & \naiveatt       & 29.2 & 96.9 & 29.2 & 175 & 29.0 & 95.3 & 29.1 & 341 \\
     & \prioratt       & 31.2 & 97.1 & 29.7 & 198 & 32.2 & 95.9 & 29.9 & 306 \\
     & \shiftatt       & 34.9 & 96.7 & 29.9 & 355 & 35.3 & 96.5 & 30.0 & 568 \\
     & \shiftaet       & 35.2 & 96.3 & 30.0 & 378 & 35.8 & 97.1 & 30.2 & 637 \\
     & \postatt        & 35.3 & 96.7 & 30.0 & 272 & 35.8 & 96.7 & 30.2 & 467 \\
     & VDBA            & 35.3 & 98.8 & 29.8 & 258 & 35.0 & 99.2 & 30.4 & 442 \\
     & \alignvdba*     & 35.4 & 99.8 & 29.8 & 280 & 35.1 & 99.3 & 30.3 & 534 \\
     & \alignvdba      & 36.1 & 98.3 & 30.1 & 268 & 35.9 & 98.8 & 30.6 & 523 \\
    \hline
    10 & No constraints & 28.3 & 77.0 & 29.7 & 113 & 26.3 & 72.4 & 29.9 & 164 \\
     & \naiveatt        & 28.9 & 97.3 & 29.1 & 145 & 29.2 & 95.3 & 29.1 & 269 \\
     & \prioratt        & 31.3 & 96.9 & 29.5 & 155 & 32.3 & 96.0 & 29.9 & 260 \\
     & \shiftatt        & 34.9 & 96.3 & 29.8 & 345 & 35.3 & 96.8 & 30.3 & 600 \\
     & \shiftaet        & 35.2 & 95.9 & 29.9 & 350 & 35.9 & 97.2 & 30.4 & 664 \\
     & \postatt         & 35.1 & 95.9 & 29.9 & 287 & 35.8 & 97.0 & 30.3 & 458 \\
     & VDBA             & 37.6 & 99.8 & 30.9 & 257 & 36.9 & 99.4 & 30.9 & 451 \\
     & Align-VDBA       & 37.5 & 99.8 & 30.9 & 353 & 37.2 & 99.5 & 31.1 & 540 \\ %
    \hline
    20 & No constraints & 28.4 & 77.2 & 29.9 & 103 & 26.3 & 72.1 & 30.0 & 177 \\
     & \naiveatt        & 28.9 & 96.9 & 29.0 & 188 & 29.1 & 95.4 & 29.3 & 325 \\
     & \prioratt        & 31.3 & 96.9 & 29.6 & 203 & 32.6 & 96.4 & 30.1 & 338 \\
     & \shiftatt        & 34.7 & 96.1 & 29.8 & 528 & 35.3 & 96.8 & 30.2 & 892 \\
     & \shiftaet        & 35.0 & 95.8 & 29.9 & 539 & 36.1 & 97.3 & 30.4 & 923 \\
     & \postatt         & 35.1 & 96.1 & 29.9 & 420 & 36.0 & 97.0 & 30.4 & 751 \\
     & VDBA             & 37.8 & 99.8 & 30.9 & 381 & 37.4 & 99.2 & 31.2 & 680 \\
     & Align-VDBA       & 37.9 & 99.8 & 30.9 & 465 & 38.0 & 99.5 & 31.3 & 818 \\
    \end{tabular}
    }
    \caption{Additional results for the real world constraints for all methods and different beam sizes. \alignvdba* denotes \alignvdba\ without alignment probability based beam allocation.}
    \label{tab:dinu-all}
\end{table}

\begin{table}[h]
\tabcolsep 3pt
    \centering
    \resizebox{\linewidth}{!}{
    \begin{tabular}{l|rrl|rrl}
 Dataset & \multicolumn{3}{c|}{IATE.414} & \multicolumn{3}{c}{Wiktionary.727} \\
 \hline
 Method & BLEU & $\mu \pm$ 95\% CI & p-value & BLEU & $\mu \pm$ 95\% CI & p-value\\
 \hline
\hline
\alignvdba & 30.1 & (30.0±1.7) &  & 30.6 & (30.6±1.2) & \\
No constraints & 29.7 & (29.7±1.7) & 0.1059 & 29.9 & (29.9±1.2) & 0.0054*\\
\naiveatt & 29.2 & (29.2±1.7) & 0.0121* & 29.1 & (29.1±1.2) & 0.0001*\\
\prioratt & 29.7 & (29.6±1.6) & 0.0829 & 29.9 & (29.8±1.2) & 0.0041*\\
\shiftatt & 29.9 & (29.8±1.6) & 0.1827 & 30.0 & (30.0±1.2) & 0.0229*\\
\shiftaet & 30.0 & (29.9±1.6) & 0.2824 & 30.2 & (30.2±1.2) & 0.0588\\
\postatt & 30.0 & (30.0±1.6) & 0.3813 & 30.2 & (30.2±1.2) & 0.0646\\
VDBA & 29.8 & (29.7±1.6) & 0.0849 & 30.4 & (30.4±1.2) & 0.0960\\
     \end{tabular}
     }
    \caption{Paired bootstrap resampling tests with 10000 bootstrap samples for BLEU on \citet{dinu-etal-2019-training} datasets for beam size 5. * denotes statistically significant difference from \alignvdba\ at power 0.05 (p-value < 0.05).}
    \label{tab:dinu-test}
\end{table}

\section{Alignment-based Token Replacement Algorithm}
\label{appendix:algo2}
The pseudocode for the algorithm used in \citet{song-etal-2020-alignment, chen-etal-2021-lexically} and our non-VDBA based methods in Section~\ref{expt:translation} is presented in Algorithm~\ref{alg:argmax}. As described in Section~\ref{sec:translation-background}, at each decoding step, if the source token having the maximum alignment at the current step lies in some constraint span, the constraint in question is decoded until completion before resuming normal decoding. 

Though different alignment methods are represented using a call to the same \textsc{Attention} function in Algorithm~\ref{alg:argmax}, these methods incur varying computational overheads. For instance, \naiveatt\ incurs little additional cost, \prioratt\ and \postatt\ involve a multi-head attention computation. For \shiftatt\ and \shiftaet, an entire decoder pass is done when \textsc{Attention} is called, thereby incurring a huge overhead as shown in Table~\ref{tab:cd}.

\begin{algorithm*}[ht]
  \caption{$k$-best extraction with argmax replacement decoding.} 
  {\bf Inputs:} A $k \times |V_T|$ matrix of scores (for all tokens up to the currently decoded ones). $k$ beam states.
  \begin{algorithmic}[1]
  \Function{Search\_Step}{beam, scores}
  \State next\_toks, next\_scores $\gets$ \Call{argmax\_k}{scores, k=2, dim=1} \Comment{Best 2 tokens for each beam}
  \State candidates $\gets$ []                                                  %
  \For{$0 \leq h < 2\cdot k$}
    \State candidate $\gets$ beam[h//2]
    \State candidate.tokens.append(next\_toks[h//2, h\%2])
    \State candidate.scores $\gets$ next\_scores[h//2, h\%2]
    \State candidates.append(candidate)
  \EndFor
  
  \State attention $\gets$ \Call{attention}{candidates}
  \State aligned\_x $\gets$ \Call{argmax}{attention, dim=1} 
  
  \For{$0 \leq h < 2\cdot k$}                                                                   %
    \If{aligned\_x[h] $\in \mcal{C}^x_i$ for some $i$ \textbf{and not} candidates[h].inprogress} \Comment{Start constraint}
        \State candidates[h].inprogress $\gets$ True
        \State candidates[h].constraintNum $\gets$ $i$
        \State candidates[h].tokenNum $\gets$ 0
    \EndIf
    \If{candidates[h].inprogress} \Comment{Replace token with constraint tokens}
        \State consNum $\gets$ candidates[h].constraintNum
        \State candidates[h].tokens[-1] $\gets$ constraints[consNum][candidates[h].tokenNum]
        \State candidates[h].tokenNum $\gets$ candidates[h].tokenNum + 1
        \If{constraints[consNum].length == candidates[h].tokenNum}
            \State candidates[h].inprogress $\gets$ False \Comment{Finish current constraint}
        \EndIf
    \EndIf
  \EndFor
  \State candidates $\gets$ \Call{remove\_duplicates}{candidates}
  \State newBeam $\gets$ \Call{top\_k}{candidates}
  \State \Return newBeam
  \EndFunction
  \end{algorithmic}
  \label{alg:argmax}
\end{algorithm*}

\section{Layer Selection for Alignment Supervision of Distant Language Pairs}

For the alignment supervision, we used alignments extracted from vanilla Transformers using the \shiftatt\ method. To do so, however, we need to choose the decoder layers from which to extract the alignments. The validation AERs can be used for this purpose but since gold validation alignments are not available, \citet{chen-etal-2020-accurate} suggest selecting the layers which have the best consistency between the alignment predictions from the two translation directions.

For the European language pairs, this turns out to be layer 3 as suggested by \citet{chen-etal-2020-accurate}. However, for the distant language pairs Hindi-English and Japanese-English, this is not the case and layer selection needs to be done. The AER between the two translation directions on the validation set, with alignments obtained from different decoder layers, are shown in Tables~\ref{tab:layer-en-hi} and~\ref{tab:layer-en-ja}.

\begin{figure*}[ht]
\begin{minipage}{0.45\textwidth}
\begin{table}[H]
    \centering
    \begin{tabular}{c|cccccc}
          & 1 & 2 & 3 & 4 & 5 & 6\\ 
        \hline
        1 & 65.5 & 55.8 & 56.1 & 95.2 & 94.6 & 96.6 \\
        2 & 59.2 & 47.5 & \textbf{44.5} & 95.1 & 91.9 & 95.8 \\
        3 & 62.6 & 52.1 & 48.3 & 93.7 & 91.4 & 95.2 \\
        4 & 88.6 & 83.3 & 82.1 & 89.9 & 88.0 & 90.3 \\
        5 & 91.6 & 87.7 & 88.5 & 91.4 & 88.8 & 90.2 \\
        6 & 93.5 & 91.1 & 92.5 & 92.5 & 90.5 & 90.7
    \end{tabular}
    \caption{AER between en$\rightarrow$hi and hi$\rightarrow$en \shiftatt\ alignments on the validation set for EnHi}
    \label{tab:layer-en-hi}
\end{table}
\end{minipage}
\hfill
\begin{minipage}{0.45\textwidth}
\begin{table}[H]
    \centering
    \begin{tabular}{c|cccccc}
          & 1 & 2 & 3 & 4 & 5 & 6 \\ 
        \hline
        1 & 93.5 & 90.0 & 94.4 & 92.2 & 95.1 & 95.1 \\ 
        2 & 86.5 & \textbf{58.7} & 86.9 & 69.4 & 87.2 & 86.2 \\ 
        3 & 87.4 & 59.4 & 87.1 & 69.1 & 87.1 & 86.2 \\ 
        4 & 89.1 & 69.1 & 85.9 & 74.2 & 84.9 & 85.4 \\ 
        5 & 93.4 & 88.5 & 89.1 & 87.1 & 86.8 & 88.1 \\ 
        6 & 93.5 & 89.4 & 90.0 & 88.1 & 87.7 & 88.7
    \end{tabular}
    \caption{AER between ja$\rightarrow$en and en$\rightarrow$ja \shiftatt\ alignments on the validation set for JaEn}
    \label{tab:layer-en-ja}
\end{table}
\end{minipage}
\end{figure*}

\end{document}